\documentclass[lettersize,journal]{IEEEtran}
\usepackage{amsmath,amsfonts}
\usepackage{algorithm}
\usepackage{algorithmicx}
\usepackage{algpseudocode}
\usepackage{amssymb}
\usepackage[colorlinks,bookmarksopen,bookmarksnumbered,citecolor=blue, linkcolor=blue, urlcolor=red]{hyperref}
\usepackage[final]{pdfpages}
\usepackage{array}
\usepackage[caption=false,font=footnotesize,labelfont=rm,textfont=rm]{subfig}
\usepackage{textcomp}
\usepackage{stfloats}
\usepackage{url}
\usepackage{verbatim}
\usepackage{graphicx}
\usepackage{cite}
\usepackage{balance}
\usepackage{algpseudocode}
\usepackage{booktabs}
\usepackage{multirow}
\usepackage{float}
\begin{document}
	\title{  
		Efficient Meta-Learning Enabled \\
		Lightweight Multiscale
		Few-Shot Object Detection in Remote Sensing Images}
	\author{Wenbin Guan, Zijiu Yang, Xiaohong Wu, Liqiong Chen, Feng Huang, Xiaohai He,   Honggang Chen
		\thanks{Wenbin Guan, Zijiu Yang, Xiaohong Wu, and Xiaohai He are with the College of Electronics and Information Engineering, Sichuan University, Chengdu 610065, China (e-mail:793763912@qq.com; yangzijiu@stu.scu.edu.cn; wxh@scu.edu.cn; hxh@scu.edu.cn).}
		\thanks{Liqiong Chen and Feng Huang are with the School of Mechanical Engineering and Automation, Fuzhou University, Fuzhou 350002, China (e-mail: liqiongchen@fzu.edu.cn; huangf@fzu.edu.cn).}
		\thanks{Honggang Chen is with the College of Electronics and Information Engineering, Sichuan University, Chengdu 610065, China, and also with the Yunnan Key Laboratory of Software Engineering, Yunnan University, Kunming 650600, China (e-mail: honggang\_chen@scu.edu.cn).}
		\thanks{Wenbin Guan and Zijiu Yang contributed equally to this work. \emph{(Corresponding author: Xiaohong Wu.)}}
		\thanks{This work has been submitted to the Elsevier for possible publication. Copyright may be transferred without notice, after which this version may no longer be accessible.}
	}

	\maketitle
	\begin{abstract}
		Presently, the task of few-shot object detection (FSOD) in remote sensing images (RSIs) has become a focal point of attention. Numerous few-shot detectors, particularly those based on two-stage detectors, face challenges when dealing with the multiscale complexities inherent in RSIs. Moreover, these detectors present impractical characteristics in real-world applications, mainly due to their unwieldy model parameters when handling large amount of data.
		In contrast, we recognize the advantages of one-stage detectors, including high detection speed and a global receptive field. Consequently, we choose the YOLOv7 one-stage detector as a baseline and subject it to a novel meta-learning training framework. This transformation allows the detector to adeptly address FSOD tasks while capitalizing on its inherent advantage of lightweight.
		Additionally, we thoroughly investigate the samples generated by the meta-learning strategy and introduce a novel meta-sampling approach to retain samples produced by our designed meta-detection head. Coupled with our devised meta-cross loss, we deliberately utilize ``negative samples" that are often overlooked to extract valuable knowledge from them. This approach serves to enhance detection accuracy and efficiently refine the overall meta-learning strategy.
		To validate the effectiveness of our proposed detector, we conducted performance comparisons with current state-of-the-art detectors using the DIOR and NWPU VHR-10.v2 datasets, yielding satisfactory results.
	\end{abstract}
	

	\section{Introduction}
	\IEEEPARstart{O}{bject} detection has become a crucial technique, playing a pivotal role in interpreting and analyzing data acquired from diverse remote sensing platforms like satellite and aerial imagery \cite{zou2023object}, \cite{huang2022evaluation}.
	In recent years, notable progress in object detection for remote sensing images (RSIs) has been propelled by convolutional neural networks (CNNs) \cite{A1}, \cite{ODRSI1}, \cite{ODRSI2}, \cite{ODRSI3}.
	Nonetheless, due to the difficulty in collecting remote sensing data, RSIs often exhibit categories with sparse labels and conventional CNN-based object detectors tend to overfit severely when confronted with limited training data. 
	Hence to make machines possess the ability to grasp new concepts from a small number of examples becomes a significant task, also known as few-shot object detection (FSOD). FSOD presents a substantial and formidable challenge for intelligent systems operating in the domain of RSI processing \cite{A2}.
	
	At present, methods for addressing FSOD can be broadly classified into two main categories: transfer learning-based and few-shot learning-based. Transfer learning involves repurposing a model initially trained on a large and diverse dataset for a specific task to a new but related task \cite{TF4}. This process typically entails adapting a pretrained model through additional training or modification to suit the requirements of the new task. Models trained on comprehensive terrestrial image datasets, for example, can be adapted for satellite imagery analysis, mitigating the necessity for extensive data collection in this domain \cite{Zhu}. The efficiency and improved performance afforded by transfer learning make it an indispensable tool in remote sensing applications. For instance, Zhao \emph{et al.} \cite{A3} devised a method that consolidates multilevel features sourced from feature pyramids. This strategy effectively truncates the information transmission pathway by implementing a bottom-up flow to disseminate localization details at lower levels. Concurrently, Wang \emph{et al.} \cite{A4} undertook the task of extracting contextual features embedded within images, aiming to discern novel objects within intricate backgrounds. 
	However, while effective, transfer learning exhibits performance degradation in cases of substantial task differences between the source and target domains. Adaptation challenges arise when significant dissimilarities exist between the two tasks, hindering the source task trained model from achieving optimal performance on the new target task. 
	
	Few-shot learning, on the other hand, focuses on the development of models that can learn from a minimal amount of labeled data. This approach is particularly salient for tasks such as the identification of rare geological features, where acquiring large datasets is often unfeasible. Few-shot learning enables models to rapidly adapt to new tasks using minimal data, which is vital in the dynamic and varied environments typical of remote sensing \cite{Finn}. For instance, Meta-YOLO \cite{B} and FRW \cite{FRW} introduced reweighting vectors to recalibrate query features across various scales. Expanding on the two-stage benchmark Meta R-CNN \cite{meta-rcnn}, Zhang \emph{et al.} \cite{Zhang} augmented the training data to handle the arbitrary orientations of objects in RSIs. SAGS-TFS \cite{A5} employed a two-way attention mechanism, fully exploiting the knowledge within support images by integrating the similarity map into the detailed embeddings of both query and support features. Nonetheless, one notable limitation of few-shot learning lies in the potential difficulty of generalizing well to unseen classes, especially when faced with diverse or highly complex target domains. The scarcity of training samples may lead to overfitting, hampering the model's ability to capture the underlying patterns and variations effectively. 
	
	In addition to the aforementioned issues, multiscale issue of objects and the vastness of data scale are also challenges brought by RSIs. Firstly, the multiscale problem of objects in RSIs arises due to the diverse scales and perspectives at which objects may appear \cite{multiscale}. Small-scale objects, possessing limited information in the image, are susceptible to being overlooked, while large-scale objects may be prone to false detections due to partial occlusion or edge effects. Many few-shot object detectors based on two-stage detectors do not explicitly address the multiscale problem, primarily due to the constraints imposed by the two-stage detection mechanism of the baseline.
	Secondly, the enormous data scale is attributed to the high resolution and extensive coverage of RSIs. 
	On one hand, there is a pressing need to address the challenges arising from the limited availability of sample data. On the other hand, efforts to expand the search for diverse datasets are being intensified \cite{largedata}. This dual emphasis necessitates that few-shot object detectors not only address the scarcity of samples from novel classes but also remain sufficiently lightweight to accommodate extremely large-scale datasets.
	Large-scale datasets are crucial for training deep learning models, but they also present computational and storage challenges. Most FSOD methods may exhibit inefficiencies in handling vast data, and the training and inference processes demand substantial hardware resources. 
	
	Taking the above-mentioned issues into consideration, this paper presents a novel few-shot object detection framework, meticulously crafted for the intricacies of RSIs. 
	Inspired by \cite{FRW}, our detector is structured into three main components: meta-feature extraction module, support set feature extraction module, and meta-detection head. Addressing the challenges posed by high-resolution and multiscale aspects in remote sensing imagery, we employ YOLOv7 \cite{yolov7} as the baseline to constitute both the meta-feature extraction and support set feature extraction modules, capitalizing on its lightweight characteristics and expansive global receptive field.
	Upon a comprehensive exploration of the meta-learning architecture, we identify valuable object knowledge embedded within the negative samples generated during meta-learning. Building upon this realization, we augment the meta-detection head with meta-sampling technique to acquire all useful output features following the channel-wise multiplication. Subsequently, we introduce the meticulously designed meta-cross loss to fortify the extraction of pertinent knowledge from negative samples, thereby enhancing data utilization in scenarios characterized by limited sample availability.
	Furthermore, based on the aforementioned revelations, we equip the detection head with the meta-cross category determination criterion, improving its capability to discern object categories in RSIs with diminished inter-class disparities. 
	We posit that an adept detector should not only achieve high detection accuracy but also exhibit a low parameter count, with due consideration given to prevailing challenges in the field during the design process.
	Thus the entirety of our module designs and combinations adheres to the principles of being lightweight and effective. 
	We name our network \emph{Lightweight Multiscale Few-Shot Detector} (LMFSODet), highlighting its effective utilization and improvement on meta-learning and the lightweight nature of the network.
	\begin{figure}
		\centering
		\includegraphics[width=4.0in, keepaspectratio]{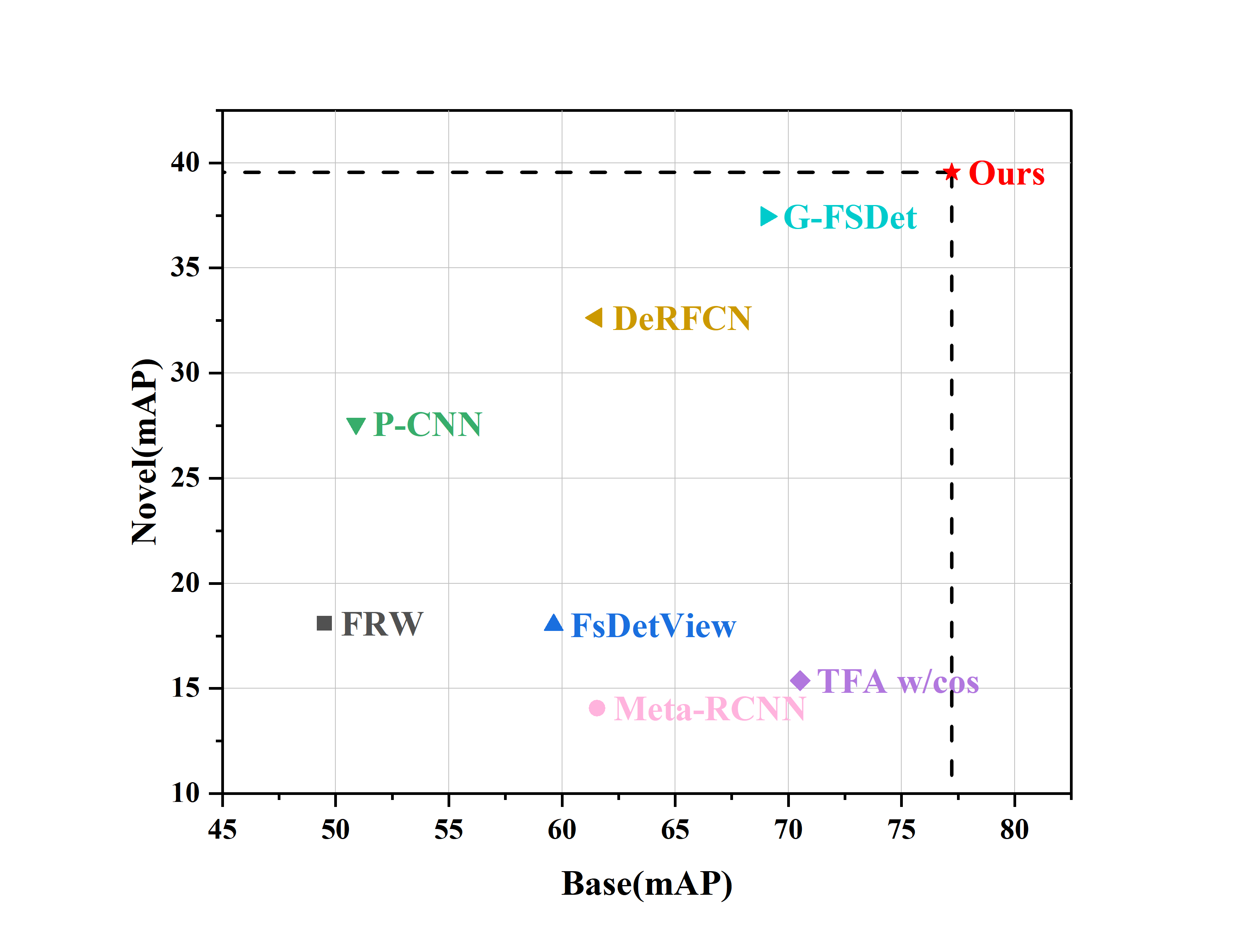}
		\caption{The comparative results of the accuracy of our proposed detector LMFSODet, against state-of-the-art detectors under the 10-shot scenario on the DIOR dataset.}
		\label{10_shot_result}
	\end{figure}
	
	This article's primary contributions come as follows.
	\begin{itemize}
		\item [1)] 
		We propose a novel object detection framework specifically tailored for RSIs, which incorporates a meta-feature extraction network, a support set feature extraction network and a meta-detection head. 
		The feature extraction segment is based on YOLOv7, and the entire network is remarkably lightweight and efficient.
		\item [2)]
		To better leverage scarce sample data, we design meta-sampling and meta-cross loss to further leverage the negative samples often overlooked. In the scenario where multiple sets of samples are generated by the feature reweighting vector, we do not singularly focus on the impact brought by a specific category like previous studies; instead, we comprehensively consider and leverage the contribution of all the samples, helping the model to learn more useful knowledge.
		\item [3)]
		In order to better discern inter-class disparities, we propose meta-cross category determination criterion, which requires identifying the object category in two dimensions.
		\item [4)]
		Our proposed detector is lightweight enough, and its detection accuracy is comparable to current advanced detectors. LMFSODet exhibits significant detection advantages on two major remote sensing few-shot datasets, with detection accuracy for both novel and base classes generally surpassing that of current state-of-the-art detectors, as shown in Fig. \ref{10_shot_result}.
	\end{itemize}
	
	\section{Related Work}
	\subsection{General Object Detection}
	For advanced general object detectors, models are primarily categorized into two distinct types based on their detection stages: two-stage and one-stage detectors.
	As for two-stage detectors, they generate region proposals–segments of the image where objects are likely to be present. Subsequently, these regions undergo a classification and bounding box refinement process. Emblematic of this category is Faster R-CNN \cite{Ren}, which employs a Region Proposal Network (RPN) to identify potential object areas, following Fast R-CNN \cite{FasterRCNN} that classifies these areas and fine-tunes their coordinates. Detectors like SPPNet\cite{SPPNet} and FPN\cite{FPN} adopt similar approaches. In contrast, one-stage detectors streamline the process by simultaneously predicting class probabilities and bounding box coordinates across the entire image, thereby eschewing a separate region proposal stage. The YOLO\cite{YOLO} series epitomizes this category, dividing the image into a grid and concurrently predicting outcomes for each cell. Similarly, the SSD\cite{SSD} leverages multiple feature maps at varying resolutions for detection purposes. CornerNet\cite{cornernet} and DETR\cite{DETR} are also classic examples of detectors that employ the one-stage detection approach. The hallmark of one-stage detectors is their speed, achieved by eliminating the region proposal phase. 
	However, one-stage detectors cannot be directly applied to the FSOD due to their reliance on substantial training data support. Acknowledging the advantages of one-stage detector's small model parameters and high detection speed, we aim to transform the chosen YOLOv7 baseline into a few-shot object detector, which leverages its detection speed advantages while addressing the scarcity of labeled data problem in RSIs.
	\subsection{Object Detection in Remote Sensing Images}
	RSIs often exhibit distinctive characteristics in terms of multiscale objects. The complexity and heterogeneity of landscapes captured in these images necessitate an exploration of objects at various scales for comprehensive analysis. This inherent multiscale nature of remote sensing imagery poses challenges and opportunities in object detection and analysis tasks.
	Currently CNNs have been widely adopted for their ability to automatically learn and extract relevant features from images, a process that is crucial for accurate detection\cite{Zhu}. And emerging trends in this field include the use of Generative Adversarial Networks (GANs) for data augmentation and enhancement, and the exploration of novel neural network architectures for improved efficiency and accuracy \cite{GOOD}.
	For instance, Tang \emph{et al.} \cite{T} develops an extension of the SSD model specifically for real-time vehicle detection in remote sensing imagery. In another approach, Liu \emph{et al.} \cite{L} modifies the SSD framework \cite{SSD} by substituting horizontal anchors with oriented ones, thereby equipping the model with the capability to detect objects at various orientation angles. 
	As for further advancements, Gao \emph{et al.} \cite{OBRM1} introduces a self-supplementary and revised anchor-free detector to enhance performance, addressing issues associated with intricate backgrounds and target scale variations. Duan \emph{et al.} \cite{OBRM2} addresses the constraints of traditional single-platform on-orbit processing by proposing a multiplatform distributed collaborative inference network. This network leverages information from multiple platforms to enhance the accuracy of inference tasks, overcoming the underutilization of multiplatform observations in the current remote sensing field. Yan \emph{et al.} \cite{OBRM3} focuses on salient object detection in RSIs and introduces the adaptive semantic network (ASNet). This novel framework, specifically designed for RSIs, tackles limitations of traditional approaches based on convolutional neural networks by offering dynamic network construction and reducing reliance on local feature extraction. 
	As stated above, the multiscale issue is an important consideration in addressing the object detection of RSIs, fully recognizing which
	our proposed LMFSODet incorporates a multiscale feature extraction strategy.
	\subsection{Few-shot Object Detection in Remote Sensing Images}
	When the core issue is the scarcity of training data, FSOD represents a burgeoning area in the field of object detection in RSIs, which aims to accurately detect objects in images using a model trained with a very small number of labeled examples per class. 
	The mainstream FSOD methods include transfer learning, few-shot learning, \emph{etc}.
	
	Transfer learning approaches have been incorporated in recent developments of FSOD detectors, with notable examples such as DH-FSDet \cite{B1}, HTD \cite{TF1}, and ADCG \cite{TF2}.
	The classic fine-tuning method of TFA \cite{TFA} is widely adopted by many detectors. The G-FSDet \cite{GFSOD}, taking into account the large intra-class diversity and low inter-class separability of geographical spatial objects, is well-suited for RSIs processing.
	Others like SAM\&BFS \cite{B2} leverage sophisticated attention mechanisms, while  TLH$^2$TD \cite{TF3} learns the spatial features of hyperspectral targets from high-resolution RSIs and achieves high-precision HTD with fused spatial–spectral features. 
	As a result, transfer learning efficiently leverages knowledge from one task to enhance performance on related tasks, proving advantageous in scenarios with sparse labeled data. 
	
	State-of-the-art FSOD detectors using a meta-learning strategy, like SAAN \cite{B4}, establish a relationship graph between Region of Interests (RoIs) and support images, integrating support information adaptively into the query feature. P-CNN \cite{PCNN} introduces a P-G region proposal network (RPN) that effectively identifies foreground objects amidst complex backgrounds under prototype guidance, garnering significant attention. FRW \cite{FRW} employs one-stage detector as the baseline, achieving satisfactory detection results. Additionally,
	Liu \emph{et al.} \cite{ML1} advocates combining a Feature Pyramid Network (FPN) and incorporating prototype features to enhance the efficacy of query features. 
	Guo \emph{et al.} \cite{ML2} proposes a novel meta-learning-based FSOD approach named DPL-Net to tackle information attenuation in RSIs. And Zhang \emph{et al.} \cite{ML3} designs ST-FSOD to address the incomplete labels problem brought by new objects.
	
	The adaptability of meta-learning enables models to learn a diverse range of tasks. However, its implementation complexity requires meticulous design and tuning. Meta-learning models may exhibit sensitivity to changes in task distributions, impacting their generalization. Additionally, the training process can be computationally intensive.
	Taking into account the aforementioned issues, we mine from samples generated through meta-learning, proposing meta-sampling and meta-cross loss, aiming to extract as much useful knowledge as possible from the given data to maximize the usage of it.
	\section{Proposed Methods}
	In this section, we will present our proposed novel approach, which encompasses the overall structure of the detector, meta-sampling technique, meta-cross loss, and meta-cross category determination criterion. Moreover, we propose a new standard for evaluating FSOD models, which better reflects the comprehensive detection performance of few-shot detectors.
	\subsection{Problem Setting}
	Similar to the previous works \cite{TFA}, \cite{GFSOD}, we adopt the standard FSOD paradigm based on meta-learning to formulate the problem. 
	Within few-shot detection scenarios, a meta-learning algorithm is trained to acquire meta-knowledge from an extensive array of detection tasks sampled from known classes, enabling effective generalization to unknown classes. Each of these sampled tasks is denoted as an episode $\mathcal{E}$, comprising a set of support images (annotated) denoted as $\mathcal{S}$ and a set of query images denoted as $\mathcal{Q}$.
	We follow \cite{B} to construct episodes from the data set of
	both seen and unseen categories. During the training phase on base classes, we exclusively employ images from the base classes for comprehensive training. Both the support set $\mathcal{S}$ and the query set $\mathcal{Q}$ are derived solely from images belonging to these base classes. Subsequently, in the training phase for novel classes, we utilize images from all classes; however, adherence to the $K$-shot requirement is paramount. For instance, in the case of $K=3$, each class undergoes training with a restricted set of three labeled samples. The specific division principles are elaborated below.
	\subsubsection{Dataset Division}
	The entire dataset, denoted as $D$, is split into two subsets: a training set $D_{TR}$ and a test set $D_{TE}$.
	\subsubsection{Class-Based Division}
	The data in the training set is further divided based on object classes. Let there be $N$ classes. Training set data is categorized as
	\begin{equation}
		D_{TR}=\left\{D_{TR,1},D_{TR,2},...,D_{TR,N}\right\},
	\end{equation}
	where $D_{TR,j}(j=0,1,2,3,...N)$ represents the data of class $i$. Similarly, the test set data is categorized as 
	\begin{equation}
		D_{TE}=\left\{D_{TE,1},D_{TE,2},...,D_{TE,N}\right\}.
	\end{equation}
	\subsubsection{Base and Novel Class Division}
\begin{figure}
		\centering
		\includegraphics[width=3.5in, keepaspectratio]{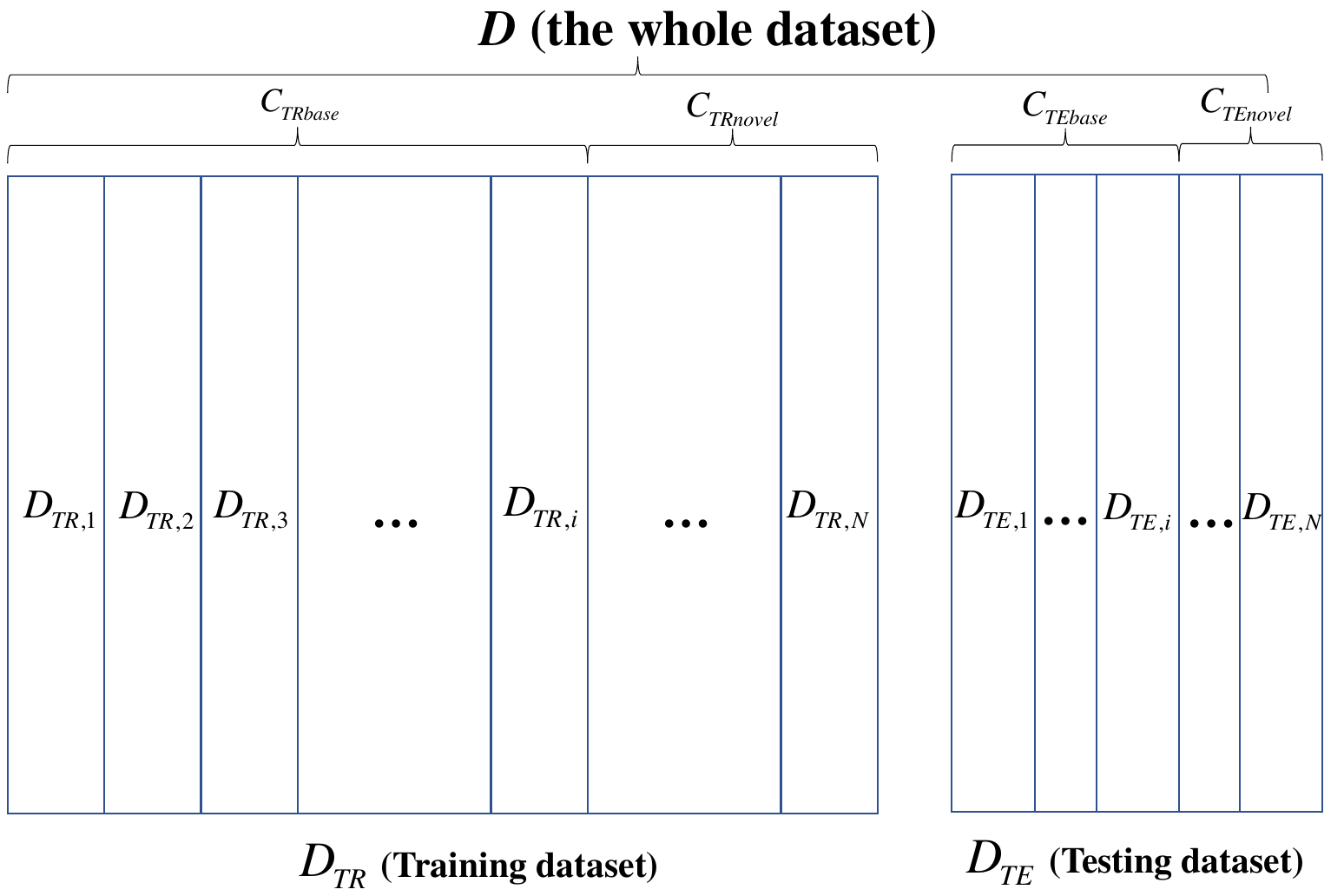}
		\caption{Dataset Partitioning Diagram. Please note that the actual number of instances for each class may vary; the illustration in the figure is for visual representation purposes only, and the sizes are not to scale.}
		\label{dataset_split}
	\end{figure}

	Following the principles from \cite{B}, assuming classes 1 to $i$ are considered base classes, and classes $i+$1  to $N$ are considered novel classes. Base class training set is denoted as 
	\begin{equation}
		C_{TRbase}=\left\{D_{TR,1},D_{TR,2},...,D_{TR,i}\right\}.
	\end{equation}
	Novel class training set is denoted as
	\begin{equation}
		C_{TRnovel}=\left\{D_{TR,i+1},...,D_{TR,N}\right\}.
	\end{equation}
	Base class test set is denoted as
	\begin{equation}
		C_{TEbase}=\left\{D_{TE,1},D_{TE,2},...,D_{TE,i}\right\}.
	\end{equation}
	Novel class test set is denoted as 
	\begin{equation}
		C_{TEnovel}=\left\{D_{TE,i+1},...,D_{TE,N}\right\}.
	\end{equation}
	The above content can be simplified and represented using Fig. \ref{dataset_split}.
	\subsubsection{Training Phases}
	As for base class training, query set $\mathcal{Q}_{1}$ during base class training is set as $\mathcal{Q}_{1}$ = $C_{TRbase}$.
	Support set $\mathcal{S}_{1}$ is formed by selecting one sample from each class in $C_{TRbase}$. While in novel class training (fine-tuning), query set $\mathcal{Q}_{2}$ is constructed based on the $K$-shot requirement, selecting $K$ samples for each class. Support set $\mathcal{S}_{2}$ is formed from $D_{TR}$, with one sample chosen from each class. It is required that
	\begin{equation}
		\mathcal{S}_{2} \in \mathcal{Q}_{2},
	\end{equation}
	ensuring the novel class training adheres to the few-shot training requirement, with each class's training data not exceeding $K$-shot. In our experiment, we adopted $K=3, 5, 10, 20.$
	\subsubsection{{Testing Phases}}
	Testing is divided into meta-training testing and meta-fine-tuning testing. In meta-training testing, $C_{TEbase}$ serves as the test set, while in meta-fine-tuning, $D_{TE}$ = ($C_{TEnovel}$ $\cup$ $C_{TEbase}$) is used.
	
	The model training process can be described as follows. Assuming the initial model is $\mathcal{M}_{i}$, after undergoing base-class training with $\mathcal{Q}_{1}$ and $\mathcal{S}_{1}$, we obtain $\mathcal{M}_{base}$. 
	Subsequently, fine-tuning is performed on $\mathcal{Q}_{2}$ and $\mathcal{S}_{2}$ to obtain the complete model. The whole process can be expressed by
	\begin{equation}
		\mathcal{M}_{i} \xrightarrow[\mathcal{Q}_{1}]{ \mathcal{S}_{1}}  \mathcal{M}_{base}\xrightarrow[\mathcal{Q}_{2}]{ \mathcal{S}_{2}}\mathcal{M}_{final}.
	\end{equation}
	
	\subsection{Overall Framework}
	\begin{figure*}[h]
		\includegraphics[width=7.0in]{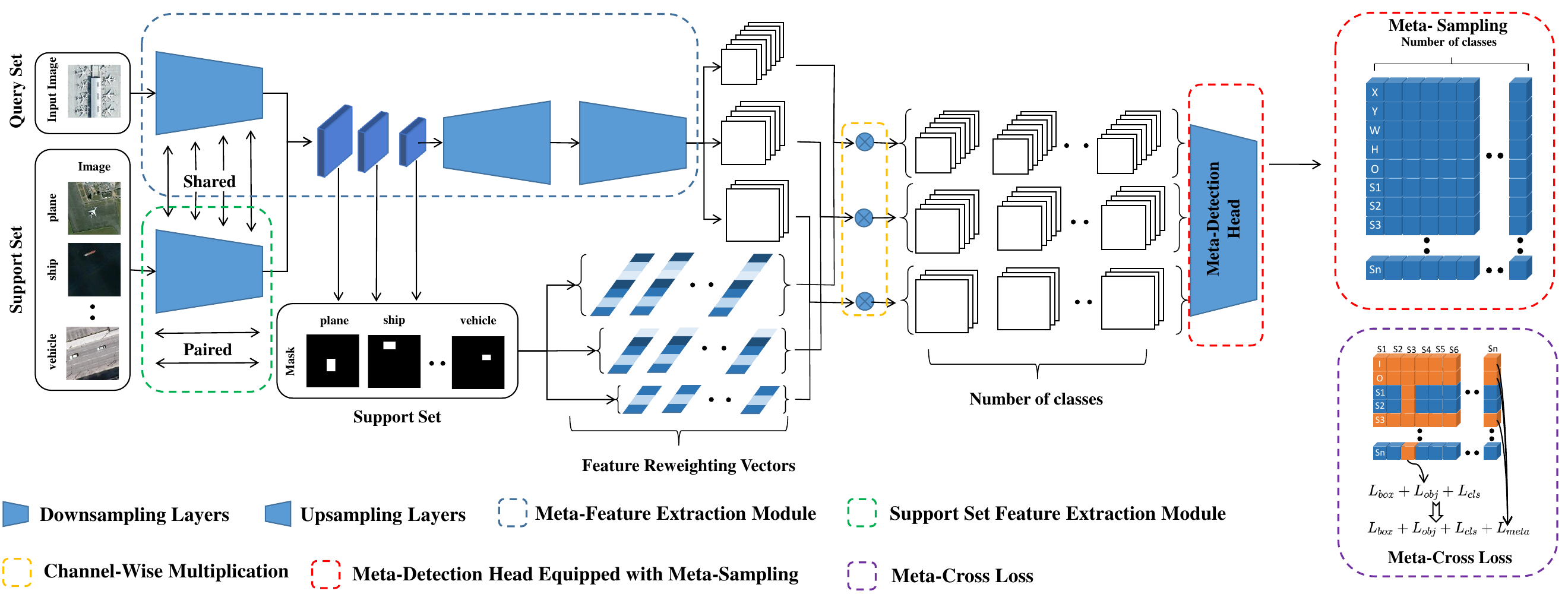}
		\centering
		\caption{Proposed meta-learning-based FSOD framework. It comprises the meta-feature extraction module, support set feature extraction module, channel-wise multiplication, meta-detection head and meta-cross loss. In contrast to mainstream meta-learning design approaches, we place the design emphasis on the post-processing of output features, specifically the meta-sampling technique and loss computation, which fully leverages the meta-learning design philosophy, pushing the advantages of meta-learning solutions for FSOD.} 
		\label{network}
	\end{figure*}
	The proposed meta-feature guided object detector framework is illustrated in Fig. \ref{network}. We opt for YOLOv7 as the baseline, leveraging its lightweight yet efficient convolutional neural network as the feature extraction module.  The feature extraction module comprises meta-feature extraction and support set feature extraction. Both segments share parameters to reduce model parameterization and ensure robust extraction of primary features. Following channel-wise multiplication of the outcomes from these feature extractions, they undergo meta-sampling via the meta-detection head. Subsequently, a meta-cross loss is applied to enhance learning from negative samples generated during meta-learning. This structured approach facilitates an integrated framework for effective object detection and meta-learning processes.
	
	\subsection{Feature Extraction Modules}
	Considering the current requirements of advanced convolutional neural networks regarding the number of input image channels \cite{CNN}, in the preceding feature extraction stage, we no longer directly input the annotated images of support set labels together with support set images. Instead, we independently perform the same feature extraction operation on the annotated images, and finally combine the two outcomes through channel-wise logical \emph{\textbf{and}} operation.

	The design of feature extraction modules in meta-learning should adhere to the principles of lightweight and effectiveness. E-ELAN \cite{yolov7} employs fundamental operations such as expansion, shuffling, and merging cardinalities to continually enhance the learning capacity of the network without disrupting the existing gradient pathways, which enables it to maintain efficient feature extraction capabilities while remaining lightweight. We contend that this methodology is highly suitable for feature extraction within the context of meta-learning frameworks, based on which we form our feature extraction modules.
	
	Following the meta-training paradigm, both the support set input and the query set input are simultaneously fed into the support set data  $\mathcal{S}\in\mathbb{R}^{w\times h\times 3}$ and query set data  $\mathcal{Q}\in\mathbb{R}^{w\times h\times 3}$.  After the downsampling layer, the data from the support set is subjected to channel-wise logical \emph{\textbf{and}} operation with the corresponding labeled image (mask image). Here, the labeled image serves as the label layer for the respective support set image and should be the result of the same operation applied to the support set image. And then a global pooling operation is applied to obtain feature reweighting vectors. The dimensions of these reweighting vectors match the number of training categories.
	Assuming the labeled image is denoted as $M\in\mathbb{R}^{w\times h\times 1}$ and the support set feature extraction operator is represented by $f_{1}(\cdot)$, then the output of the support set is
	\begin{equation}
		\mathcal{O}_{1} = f_{1}(M,\mathcal{S}) \in \mathbb{R}^{1\times 1\times c},
	\end{equation}
	where $c$ represents the number of channels.
	
	For the query set, meta-features are obtained through detailed downsampling and upsampling layers. Suppose the meta-feature extraction operator is represented by $f_{2}(\cdot)$, then the output is denoted by
	\begin{equation}
		\mathcal{O}_{2} = f_{2}(\mathcal{Q}) \in \mathbb{R}^{w\times h\times c}.
	\end{equation}
	Finally, features obtained from the support set and the query set undergo channel-wise multiplication to generate the output features. The dimensions of these output features should align with the number of training categories, which can be expressed by
	\begin{equation}
		\mathcal{O}= \mathcal{O}_{1} \otimes \mathcal{O}_{2},
	\end{equation}
	where $\otimes$ denotes channel-wise multiplication.
	$\mathcal{O}$ represents feature maps at three scales and can be described as
	\begin{equation}
		\mathcal{O}= \lbrace w_{ji}\times h_{ji}\times c_j\rbrace,
	\end{equation}
	where $j=1,2,3,i=1,2,3,...,N, N$ represents the catogory number in training, and $w_{ji},h_{ji}, c_j$ represent the size and number of channels of the image.
	Due to the model's extraction and computation of features from three different scales, there are three sets of feature maps, further 
	mitigating the challenge of large-scale variations in the spatial extent of RSIs.
	
	\subsection{Meta-Sampling Strategy}
	
\begin{figure}
		\centering
		\includegraphics[width=3.5in, keepaspectratio]{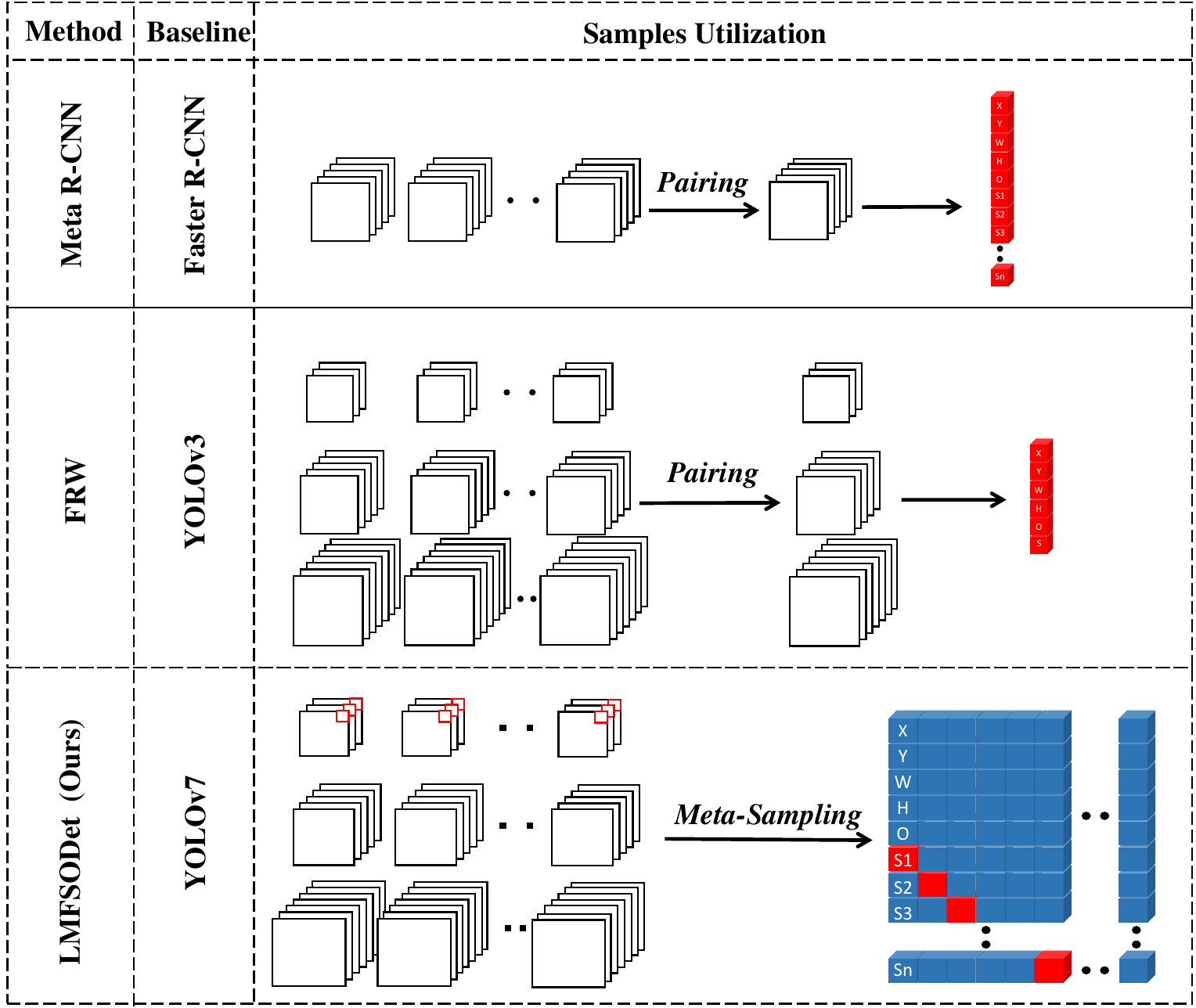}\\
		\caption{Comparison of samples handling patterns among Meta-RCNN \cite{meta-rcnn}, FRW \cite{FRW}, and the network we propose (LMFSODet).}
		\label{meta_compa}
	\end{figure}

	One of the most significant differences between our designed detector and many other object detectors employing meta-learning strategies lies in our meta-sampling approach. Some previous studies reckon that only the component of the feature reweighting vectors in the support set matching the detected object category is meaningful,
	building upon which we find that the results of other groups still contain useful knowledge. 
	Hence we retain all the useful data and subject it to subsequent meta-cross loss analysis, allowing the network more opportunities to learn useful knowledge, which provides a new perspective for improving meta-learning strategies. 
	As shown in Fig. \ref{meta_compa}, both Meta R-CNN \cite{meta-rcnn} based on Faster R-CNN \cite{Ren} and FRW \cite{FRW} based on YOLOv3 \cite{yolov3} with feature extraction of three scales choose the group matching the feature reweighting vectors for loss computation. Obviously these detectors only use one group of support set samples that indeed have the most significant impact on the detection results, and we refer to them as meta-positive samples $\mathcal{P}_{meta}$. However, valuable knowledge is also present in the other groups of samples introduced by the feature reweighting vectors, and we term such samples as meta-negative samples $\mathcal{N}_{meta}$. Our meta-sampling approach involves sampling from both $\mathcal{P}_{meta}$ and $\mathcal{N}_{meta}$, thereby maximizing the utilization of the available data and improvement on meta-learning.
	In a certain sense, this approach shares similarities with data augmentation, as the introduced samples can be considered as extra data. However, it fundamentally differs from data augmentation because it searches for more meaningful training samples instead of simply manipulating the data itself to increase the mount of the training data.
	
	The process illustrated in the last row of Fig. \ref{meta_compa} represents the meta-sampling procedure. The objects in the query set, multiplied by channel-wise multiplication with the support set, and passed through the detection layer, result in a group of vector of objects and scores equal to the number of categories in the support set. The output of the head could be expressed as
	\begin{equation}
		\mathcal{O}_{head}=\lbrace x_i,y_i,w_i,h_i,o_i,s_{1i},...,s_{Ni}\rbrace = S_{meta}(\mathcal{O}),
	\end{equation}
	where $\lbrace x_i,y_i,w_i,h_i\rbrace$ represents the prediction box of the detected object, 
	$o_i$ denotes the confidence scores for objects within the prediction boxes, and 
	$\lbrace s_{1i},s_{2i}...s_{Ni}\rbrace$ indicates the probability for each classification. As for other meta-learning-based detectors, they choose a constant $i$, so only one group is selected and $\mathcal{P}_{meta}$ is used.
	While in our detector, we utilize $i=1,2,...,N$, so all useful samples are retained, which contain $\mathcal{P}_{meta}$ and $\mathcal{N}_{meta}$.
	\subsection{Meta-Cross Loss}
	\begin{figure*}[h]
		\includegraphics[width=7.0in]{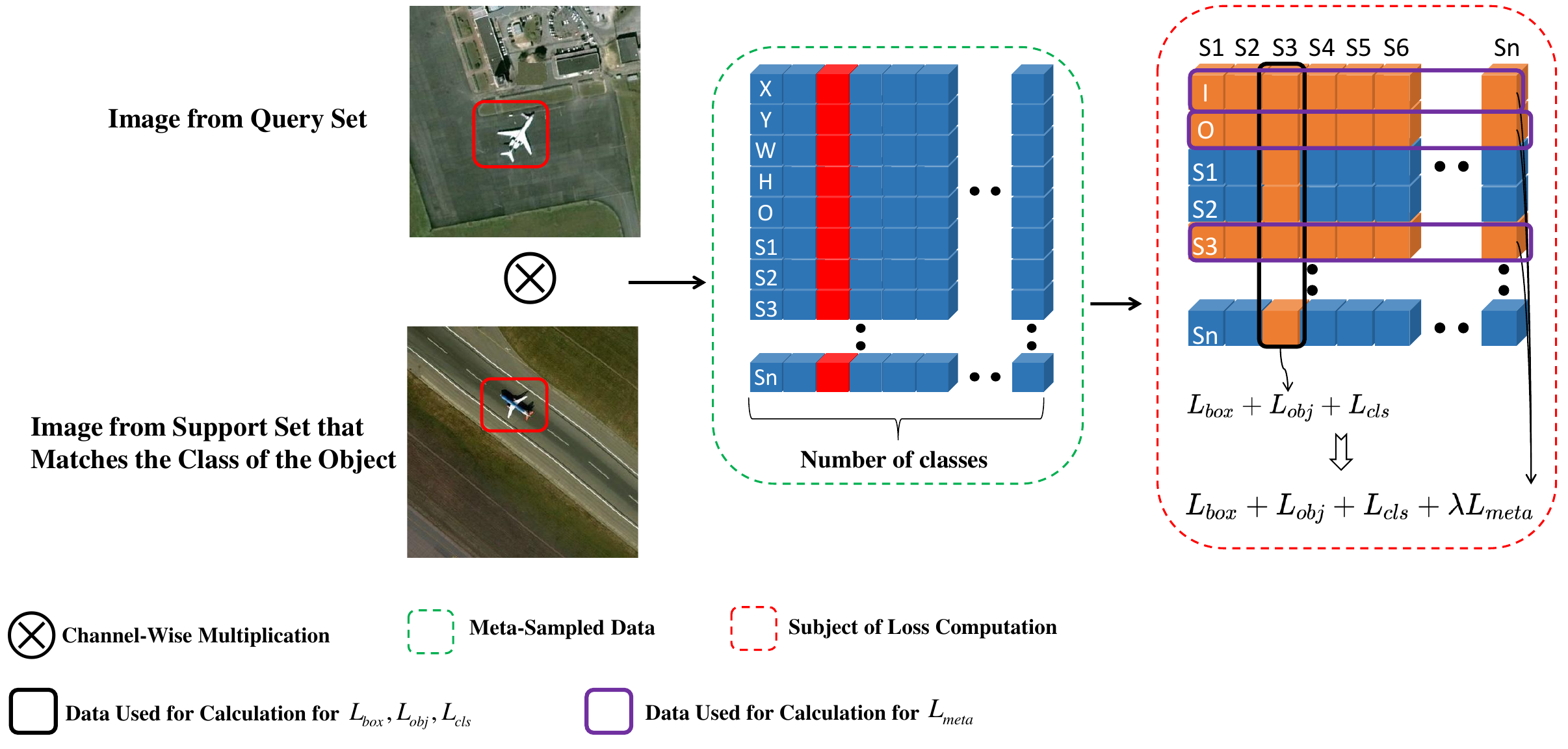}
		\centering
		\caption{The illustration of the loss computation process. The loss comprises four components, namely the coordinate loss, object confidence loss, classification loss and meta-cross loss, each governed by the corresponding principles illustrated in the rightmost diagram. Leveraging the benefits of meta-sampling, we conduct loss computation utilizing the informative samples, leading to a significant enhancement in sample utilization efficiency. } 
		\label{LOSS_whole}
	\end{figure*}
	In the context of mainstream loss algorithms, we incorporate the coordinate loss $\mathcal{L}_{box}$, object confidence loss $\mathcal{L}_{obj}$, classification loss $\mathcal{L}_{cls}$, and meta-cross loss $\mathcal{L}_{meta}$. 
	And our whole loss can be expressed by
	\begin{equation}
		\label{loss_compu}
		\mathcal{L}=\mathcal{L}_{box}+\mathcal{L}_{obj}+\mathcal{L}_{cls}+\lambda\mathcal{L}_{meta},
	\end{equation}
	where $\lambda$ represents the weight value of $\mathcal{L}_{meta}$, which will be discussed in the subsequent section.
	In contrast to the loss handling approaches of previous studies, Eq. \textcolor{blue}{(\ref{loss_compu})} introduces $\mathcal{L}_{meta}$ during loss computation, assigning it a fixed weight $\lambda$, which better utilizes $\mathcal{N}_{meta}$ produced by channel-wise multiplication. 
	\begin{algorithm}
		\caption{The steps for training and testing}
		\label{Alg1}
		\begin{algorithmic}[1]
			\Require Metafeature object detection framework $f_{\theta}$, Meta Positive sample $\mathcal{P}_{meta}$, Meta Negative sample $\mathcal{N}_{meta}$
			\State Randomly initialize the $M_{base}$ parameters $\theta _{base}$.
			\For {each training batch $\mathcal{Q} \in D_{base},\mathcal{S} \in D_{base}$}
			\State  $\mathcal{O}=f_{\theta _{base}}(\mathcal{Q},\mathcal{S})$
			\State Generate Meta sample $\mathcal{S}_{meta}$ form Output $\mathcal{O}$, $\mathcal{P}_{meta} \in \mathcal{S}_{meta}, \mathcal{N}_{meta} \in \mathcal{S}_{meta}$.
			\State Calculate the whole loss $\mathcal{L} =\mathcal{L}_{box}+\mathcal{L}_{obj}+\mathcal{L}_{cls}+\lambda\mathcal{L}_{meta}$ using $\mathcal{P}_{meta}$ and $\mathcal{N}_{meta}$.
			\State Update $M_{base}$ parameters to $\theta _{base} =\theta _{base}-\eta \bigtriangledown_{\theta _{base}} \mathcal{L} $.
			\EndFor
			\State Inherit the $M_{k-shot}$ parameters $\theta _{k-shot}$ form the $M_{base}$ parameters $\theta _{base}$.
			\State Divide $\theta _{k-shot}$ into two parts: Frozen parameters $\widetilde{\theta}_{k-shot}$ and Unfrozen parameters $\widetilde{\widetilde{\theta}}_{k-shot}$
			\For {each training batch $\mathcal{Q} \in D_{k-shot},\mathcal{S} \in D_{k-shot}$}
			\State  $\mathcal{O}=f_{\widetilde{\widetilde{\theta}}_{k-shot}}(f_{\widetilde{\theta}_{k-shot}}(\mathcal{Q},\mathcal{S}))$
			\State Generate Meta sample $\mathcal{S}_{meta}$ form Output $\mathcal{O}$, $\mathcal{P}_{meta} \in \mathcal{S}_{meta}, \mathcal{N}_{meta} \in \mathcal{S}_{meta}$.
			\State Calculate the whole loss $\mathcal{L} =\mathcal{L}_{box}+\mathcal{L}_{obj}+\mathcal{L}_{cls}+\lambda\mathcal{L}_{meta}$ using $\mathcal{P}_{meta}$ and $\mathcal{N}_{meta}$.
			\State Update $M_{k-shot}$ parameters to $\widetilde{\widetilde{\theta}}_{k-shot} =\widetilde{\widetilde{\theta}}_{k-shot}-\eta \bigtriangledown_{\widetilde{\widetilde{\theta}}_{k-shot}} \mathcal{L} $.
			\EndFor
			\For {each testing batch $\mathcal{Q} \in D_{novel},\mathcal{S} \in D_{k-shot}$}
			\State $\mathcal{O}=f_{\theta_{k-shot}}(\mathcal{Q},\mathcal{S})$
			\State Generate predicted bounding boxes and classes from $\mathcal{O}$ by using the meta-detection head.
			\EndFor
		\end{algorithmic}
	\end{algorithm}
	Assuming that the training image contains an airplane object, with the airplane class of interest concentrated in the third category in the support set, our loss region of interest is illustrated in the orange portion as shown in Fig. \ref{LOSS_whole}. 
	Assuming the loss transformation operator $Z(\cdot)$, the result on the rightmost side of Fig. \ref{LOSS_whole} can be expressed as
	\begin{equation}
		L_i,O_i,S_{ij}=Z(\mathcal{O}_{head}),
	\end{equation}
	where $L_i,O_i$ respectively represent the data utilized for regression loss computation and object confidence loss computation.
	Due to the effect of meta-sampling, these data are no longer one-dimensional but rather $N$-dimensional, where $N$ is the number of classes in the support set, $i=1,2,...,N.$ As for the data utilized for classification loss $S_{ij}$, it transforms into an $N \times N$ matrix form accordingly, thus $j=1,2,...,N$. 
	
	Assuming the corresponding loss operators of  $\mathcal{L}_{box}$, $\mathcal{L}_{obj}$, and $\mathcal{L}_{cls}$ are $M_1(\cdot), M_2(\cdot), M_3(\cdot)$, 
	based on \cite{yolov7}, there should be
	\begin{equation}
		\begin{split}
			\mathcal{L}_{box}= M_1(L_3),\\
			\mathcal{L}_{obj}= M_2(O_3),\\
			\mathcal{L}_{cls}= M_3(S_{i3}) ,
		\end{split}
	\end{equation}
	where $i=1,2,3,...,N.$
	The above content can be demonstrated in Fig. \ref{LOSS_whole}.
	As for $\mathcal{L}_{meta}$, we adopt the Cross-Entropy (CE) loss, emphasizing the effective penalization and minimization of the difference between predicted probabilities and actual class labels. Given the logarithmic nature of the loss function, it places increased emphasis on confident errors in predictions. Integrating CE loss as the primary computational framework for meta-cross loss facilitates a more precise model training within the meta-learning framework. This strategic choice assists us in achieving heightened sensitivity to class prediction accuracy and leveraging the emphasis on confident errors inherent in CE loss. Consequently, employing CE loss as the cornerstone of the meta-cross loss contributes to the refinement of model training under the meta-learning paradigm.
	Considering the loss region of interest shown in Fig. \ref{LOSS_whole}, the input first undergoes an activation process, as expressed by
	\begin{equation}
		\begin{split}
			I_{meta1}=SoftMax(L_i).\\
			I_{meta2}=SoftMax(O_i).\\
			I_{meta3}=SoftMax(S_{3j}).
		\end{split}
	\end{equation}
	And then calculate the three components of 
	$\mathcal{L}_{meta}$, which are
	\begin{equation}
		\begin{split}
			L_{meta1}=-\frac{1}{N}\sum_{i=1}^{N}one\_hot_{yi}\times log(I_{meta1}).\\
			L_{meta2}=-\frac{1}{N}\sum_{i=1}^{N}one\_hot_{yi}\times log(I_{meta2}).\\
			L_{meta3}=-\frac{1}{N}\sum_{i=1}^{N}one\_hot_{yi}\times log(I_{meta3}),
		\end{split}
	\end{equation}
	where $one\_hot_{yi}$ represents the one-hot encoding of the ground truth values. In our assuming that the object belongs to the third class in the support set,
	there should be 
	\begin{equation}
		one\_hot_{yi}=[0,0,1,0...0].
	\end{equation}
	Thus the whole $\mathcal{L}_{meta}$ is calculated by
	\begin{equation}
		\mathcal{L}_{meta}=	L_{meta1}+	L_{meta2}+	L_{meta3}.
	\end{equation}
	The overall algorithm can be pseudocoded as Algorithm \ref{Alg1}. 
	Firstly, we train a base model with randomly initialized parameters, utilizing meta-sampling techniques and calculating the whole loss. Then we emphasize the inheritance of parameters to the $K$-shot model, where parameter division into frozen and unfrozen parts occurs. During the $K$-shot model training, we further refine the parameters using a similar set of steps. The testing phase involves predicting bounding boxes and classes for novel data, leveraging our proposed meta-learning techniques to enhance discrimination between novel and base classes. With the integration of all modules we design, object detection performance is optimized.
	\subsection{Meta-Cross Prediction Method}
	\begin{figure*}
		\includegraphics[width=6.0in]{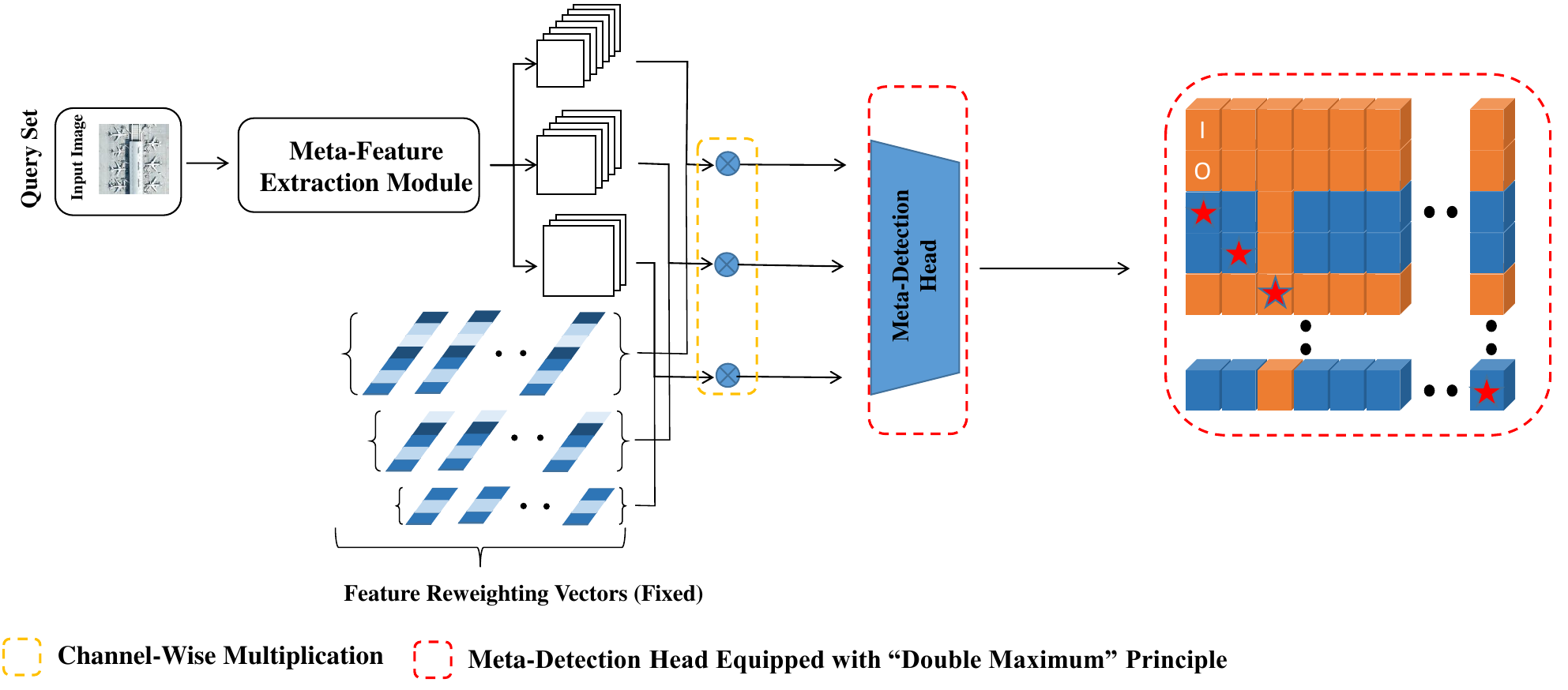}
		\centering
		\caption{Schematic diagram of the mechanism of LMFSODet prediction mechanism. The input image is fed into the query set input end, as indicated in the previous training process. It undergoes channel-wise multiplication with the reweighting vector obtained after comprehensive training to generate the output. Inference is performed based on the dual-max principle.}
		\label{predict_mechanism}
	\end{figure*}
	Based on meta-sampling and meta-cross loss, we propose the meta-cross category determination criterion to strengthen inter-class discrimination during prediction. The whole process can be illustrated in Fig. \ref{predict_mechanism}.
	
	Considering that the parameters of the reweighting module layers are relatively fixed, the reweighting part will be involved in prediction in a constant manner, thereby reducing the mount of parameters. After computing the output features,
	they similarly pass through the detection head. In the prediction phase, the results are analyzed for category determination based on the scores they form. During the category determination, there will be a two-dimensional probability score grid as shown on the rightmost side of Fig. \ref{predict_mechanism}. Let the prediction operator be $A(\cdot)$, and the location, object and classification scores  can be respectively represented as
	\begin{equation}
		LI_{i},OI_{i},C_{ij} = A(\mathcal{O}_{head}),
	\end{equation}
	where $i$ represents the category sequence in the support set, $i=1,2,3,...,N$ and $j$ represents the probability for each class, $j=1,2,3,...,N$. We propose the ``Double Maximum" Principle (DMP)  for category determination, which can be described as
	\begin{equation}
		\label{DM}
		\begin{split}
			C_{oo} = Max(C_{o1},C_{o2},C_{o3},...,C_{oN}),\\
			C_{oo} = Max(C_{1o},C_{2o},C_{3o},...,C_{No}),
		\end{split}
	\end{equation}
	where $C_{oo}$ denotes the correct category.
	To be precise, the correct category can only be determined when the positions marked with asterisks in Fig. \ref{predict_mechanism} satisfy DMP or Eq. \textcolor{blue}{(\ref{DM})}.
	
	The whole process is stated in Algorithm \ref{Alg2}.
	In detail, if the input image contains a object, the object will be processed by the feature extraction modules and multiplied by the feature reweighting vectors to obtain output features. Subsequently, after passing through the prediction layer, the meta-detection head identifies appropriate predictions based on DMP and a confidence threshold. In practice, this has been verified and found to be accurate. 
	\begin{algorithm}[h]
		\caption{The steps for predicting}
		\label{Alg2}
		\begin{algorithmic}[1]
			\Require Trained model $f_{\theta_{prediction}}$.
			\State Feed images on the query input.
			\State Compute the feature output through reweighting vectors trained and determined before.
			\State $\mathcal{O}=f_{\theta_{prediction}}(\mathcal{Q},\mathcal{S})(\mathcal{S}$ viewed as constant)
			\State Feature output passes through the meta-detection head.
			\State Follow DMP to select class and predict objects.
			\State Objects predicted.
			
		\end{algorithmic}
	\end{algorithm}
	\subsection{Enhanced Class-Weighted Evaluation Standard (ECES)}
	Most studies now utilize the mean Average Precision (mAP) with an IoU threshold set at 0.5 to assess the efficacy of few-shot object detectors \cite{MAP}. The mAP represents the average of Average Precision (AP) across all categories. A higher mAP value signifies superior performance. The AP metric is computed by averaging the Precision (P) values over the Recall (R) range from 0 to 1, under a specific IoU threshold within a single category. This calculation can be expressed as follows.
	\begin{equation}
		\label{MAP}
		AP=\sum_{R\epsilon(0,0,...,1,1)}^{} \frac{1}{11} P(R).
	\end{equation}
	
	For the performance evaluation of few-shot object detectors, the majority of studies now include the detection accuracy of both base and novel classes in the assessment scope, employing the average accuracy across all classes as the comprehensive metric for evaluating the performance of few-shot object detectors. It is important to note that the training samples for novel classes are much smaller than those for base classes. 
	However, in general few-shot datasets, the number of novel class categories is much lower than the base classes. If each class is treated equally for overall accuracy, it would be unfair for the factor of novel classes, lacking comprehensiveness and fairness in overall performance assessment.
	
	\begin{table*}[htb] 	
		\caption{Four different novel/base split settings on the DIOR dataset. }
		\label{tab1}
		\centering
		\begin{tabular}{c|ccccc|c} 	
			\toprule  			
			split & &  & Novel  &  &  & Base \\
			\midrule  			
			1 & Baseball field & Basketball court & Bridge & Chimney & Ship & Rest \\
			2 & Airplane & Airport & Expressway toll station & Harbor & Ground track field & Rest\\
			3 & Dam & Golf course & Storage tank & Tennis court & Vehicle & Rest \\
			4 & Express service area & Overpass & Stadium & Train station & Windmill & Rest \\
			\bottomrule  		
		\end{tabular}
	\end{table*}
	
	Therefore, we propose a new evaluation standard, named Enhanced Class-Weighted Evaluation Standard (ECES). Assuming the number of base class categories in the dataset is $B$, with a base class detection accuracy of $B_{APi},i=1,2,3,...,B$ and base weight $W_B$, and the number of novel class categories is $N$, with a novel class detection accuracy of $N_{APj},j=1,2,3,...,N$ and novel weight $W_N$, its calculation formula for the comprehensive performance accuracy $EM_{AP}$ is provided by
	\begin{equation}
		\label{ECES}
		EM_{AP} = \frac{\sum_{i=1}^{B} B_{APi}\times W_B  + \sum_{j=1}^{N}N_{APj}\times W_N}{ B\times W_B+ N\times W_N},
	\end{equation}
	where $W_B$ and $W_N$ should satisfy
	\begin{equation}
		\label{ECES_weight}
		B\times W_B = N\times W_N.
	\end{equation}
	This criterion assigns equal weight to novel classes and base classes, as we believe it aligns more with the fundamental task of FSOD. 
	\section{Experiments and results}
	In this section, we compare the performance of our detector with state-of-the-art few-shot detectors on two major remote sensing datasets DIOR and NWPU VHR-10.v2. Additionally, we conduct ablation analysis on our proposed methods and modules. 
	\subsection{Datasets}
	\subsubsection{DIOR dataset}
	The DIOR dataset \cite{DIOR} signifies a significant advancement in earth observation, particularly in object detection. It overcomes limitations of prior datasets, standing out for its vast scale, diversity, and complexity. Comprising 23,463 high-resolution images, totaling 192,472 object instances across 20 categories, DIOR maintains uniformity with 800×800 pixels for all images. Spatial resolutions vary from 0.5m to 30m, capturing the diversity in object sizes resulting from different sensors and natural disparities. As a preferred choice in remote sensing research, DIOR serves as a benchmark for detector performance, a role we also adopt in our proposed detector evaluation.
	\subsubsection{NWPU VHR-10.v2 dataset}
	
	Derived from NWPU VHR-10, NWPU VHR-10.v2 \cite{NWPUv2} undergoes slight modifications. Li \emph{et al.} \cite{NWPUv2} standardizes the NWPU VHR-10 dataset, curating 1172 images resized to 400 × 400 pixels. These images are selectively cropped from the varied original positive image set, maintaining coverage of ten geospatial object classes. Despite adaptations, NWPU VHR-10.v2 remains a challenging dataset, featuring airplanes, ships, storage tanks, baseball diamonds, tennis and basketball courts, ground track fields, harbors, bridges, and vehicles. Regarded by researchers, NWPU VHR-10.v2 was employed to assess our proposed detector's performance.
	\subsection{Experimental Settings}
	Following the criteria outlined in \cite{B} and the widely adopted guidelines for evaluating detector performance based on novel class/base class partitioning, we partitioned the two datasets in accordance with the specifications detailed in Tables \ref{tab1} and \ref{tab2}.
	\begin{table}[tb] 	
		\caption{Two different novel/base split settings on the NWPU-VHR10.v2 dataset.}
		\label{tab2}
		\centering
		\begin{tabular}{c|ccc|c} 	
			\toprule  			
			split &  & Novel &  & Base \\
			\midrule  			
			1 & Airplane & Baseball diamond & Tennis court & Rest \\
			2 & Basketball & Ground track field & Vehicle & Rest \\
			\bottomrule  		
		\end{tabular}
	\end{table}
	\begin{table*}[htb] 	
		\caption{
			The performance of the state-of-the-art FSOD detectors is evaluated on the NWPU VHR-10.v2 test set across different settings, including 3-shot, 5-shot, 10-shot, and 20-shot scenarios. The best results are indicated in red, while the second-best results are highlighted in blue. All data results are obtained by averaging the results of three independent experiments.}
		\label{tab3}
		\renewcommand{\arraystretch}{1.2}
		\centering
		\begin{tabular}{cc|ccc|ccc|ccc|ccc} 	
			\toprule  			
			Split & Method &  & 3-shot &  & & 5-shot  &  & &  10-shot &  &  &20-shot  &  \\
			\midrule  			
			&  & Base & Novel & All & Base & Novel & All & Base & Novel & All & Base & Novel & All \\
			\multirow{8}*{1}
			& FRW \cite{FRW} & 83.13 & 15.35 & 62.80 & 82.78 & 16.24 & 62.82 & 83.89 & 24.00 & 65.92 & 82.80 & 27.16 & 66.11 \\
			
			& Meta-RCNN \cite{meta-rcnn} & 87.00 & 20.51 & 67.05 & 85.74 & 21.77 & 66.55 & 87.01 & 26.98 & 69.00 & 87.29 & 28.24 & 69.57 \\
			
			& FsDetView \cite{FsD} & 87.68 & 24.56 & 68.75 & 87.77 & 29.55 & 70.31 & 87.75 & 31.77 & 70.96 & 87.83 & 32.73 & 71.30 \\
			
			& P-CNN \cite{PCNN} & 82.84 & 41.80 & 70.53 & 82.89 & 49.17 & 72.79 & 83.05 & 63.29 & 78.11 & 83.59 & 66.83 & 78.55 \\
			
			& TFA w/cos \cite{TFA} & \textcolor{red}{\textbf{89.35}} & 8.80 & 65.19 & \textcolor{blue}{\textbf{89.60}} & 9.49 & 64.65 & \textcolor{blue}{\textbf{89.95}} & 9.26 & 65.74 & 89.62 & 10.83 & 65.98 \\
			
			& DeRFCN \cite{DeRCFN} & 87.22 & 37.90 & 73.33 & 86.87 & 46.08 & 74.63 & 87.29 & 62.95 & 79.99 & 87.26 & 64.61 & 80.46 \\
			
			& G-FSDet \cite{GFSOD} & \textcolor{blue}{\textbf{89.11}} & \textcolor{blue}{\textbf{49.05}} & \textcolor{red}{\textbf{77.01}} & 88.37 & \textcolor{blue}{\textbf{56.10}} &\textcolor{blue}{\textbf{78.64}} & 88.40 & \textcolor{blue}{\textbf{71.82}} & \textcolor{blue}{\textbf{83.43}} & \textcolor{blue}{\textbf{89.73}} & \textcolor{blue}{\textbf{75.41}} & \textcolor{blue}{\textbf{85.44}} \\
			
			& Ours & 85.15 & \textcolor{red}{\textbf{53.16}} & \textcolor{blue}{\textbf{75.55}} & \textcolor{red}{\textbf{90.62}}  & \textcolor{red}{\textbf{72.65}} & \textcolor{red}{\textbf{85.23}} & \textcolor{red}{\textbf{91.16}} & \textcolor{red}{\textbf{87.97}} & \textcolor{red}{\textbf{90.20}} & \textcolor{red}{\textbf{91.67}} &\textcolor{red}{\textbf{90.20}} & \textcolor{red}{\textbf{91.23}} \\
			\midrule  
			\multirow{8}*{2} & FRW \cite{FRW} & 85.34 & 28.61 & 68.32 & 85.75 & 31.45 & 69.45 & 85.49 & 32.53 & 69.60 & 86.58 & 33.93 & 70.79 \\
			
			& Meta-RCNN \cite{meta-rcnn}& 86.86 & 21.41 & 67.23 & 87.38 & 35.34 & 71.77 & 87.56 & 37.14 & 72.43 & 87.26 & 39.47 & 72.92 \\
			
			& FsDetView \cite{FsD}& 88.11 & 39.01 & 73.38 & 89.34 & 40.31 & 74.63 & 89.34 & 45.09 & 76.07 & 89.31 & 46.28 & 76.40 \\
			
			& P-CNN \cite{PCNN}&  81.03 & 39.32 & 68.52 & 81.18 & 46.10 & 70.70 & 80.93 & 55.90 & 73.41 & 81.21 & 58.37 & 75.50 \\
			
			& TFA w/cos \cite{TFA}& \textcolor{blue}{\textbf{90.14}} & 11.14 & 66.44 & \textcolor{red}{\textbf{91.19}}& 12.46 & 67.57  &\textcolor{blue}{\textbf{90.79}} & 11.35 & 66.96 & 90.37 & 11.56 & 66.73 \\
			
			& DeRFCN \cite{DeRCFN}& 88.47 & 39.19 & 73.69 & 88.17 & 45.56 & 75.29 & 88.25 & 54.05 & 77.99 & 88.05 & 57.38 & 78.83 \\
			
			& G-FSDet \cite{GFSOD}& 89.99 & \textcolor{red}{\textbf{50.09}} & \textcolor{red}{\textbf{78.02}} & \textcolor{blue}{\textbf{90.52}} &  \textcolor{red}{\textbf{58.75}} & \textcolor{red}{\textbf{80.99}} & 89.23 & \textcolor{blue}{\textbf{67.00}} & \textcolor{blue}{\textbf{82.56}} & \textcolor{blue}{\textbf{90.61}} & \textcolor{blue}{\textbf{75.86}} & \textcolor{blue}{\textbf{86.13}} \\
			
			& Ours & \textcolor{red}{\textbf{91.11}} & \textcolor{blue}{\textbf{39.72}} & \textcolor{blue}{\textbf{75.69}} & 89.92 &\textcolor{blue}{\textbf{57.57}} & \textcolor{blue}{\textbf{80.22}} &\textcolor{red}{\textbf{91.89}}  & \textcolor{red}{\textbf{73.08}}& \textcolor{red}{\textbf{86.25}} & \textcolor{red}{\textbf{93.91}} &\textcolor{red}{\textbf{82.37}} & \textcolor{red}{\textbf{90.45}}\\
			\bottomrule  		
		\end{tabular}
	\end{table*}
	\begin{figure*}[htb]
		\centering
		
		\subfloat[Results of G-FSDet \cite{GFSOD} on the NWPU VHR-10.v2 dataset in the 10-shot scenario.]{\includegraphics[width=7.0in]{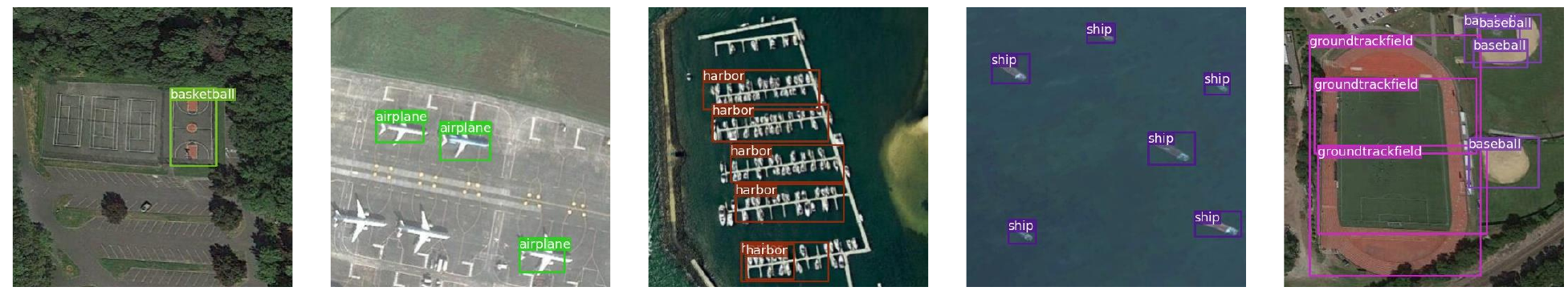} }
		
		\subfloat[Results of LMFSODet (ours) on the NWPU VHR-10.v2 dataset in the 10-shot scenario.]{\includegraphics[width=7.0in]{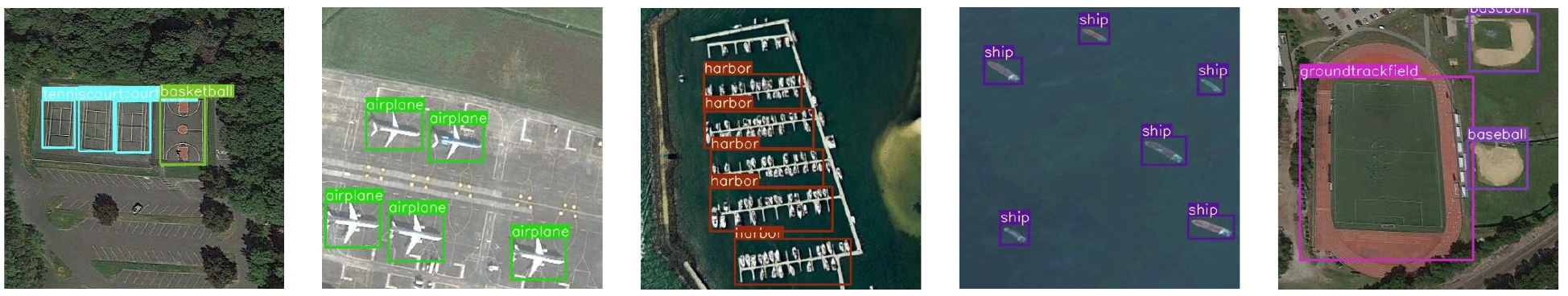} }
		\caption{Visualization results of LMFSODet (ours) compared to G-FSDet \cite{GFSOD} on the NWPU VHR-10.v2 dataset in the 10-shot scenario.}
		\label{NWPU_VISU}
	\end{figure*}
	\begin{figure}[htb]
		\centering
		\includegraphics[width=3.6in, keepaspectratio]{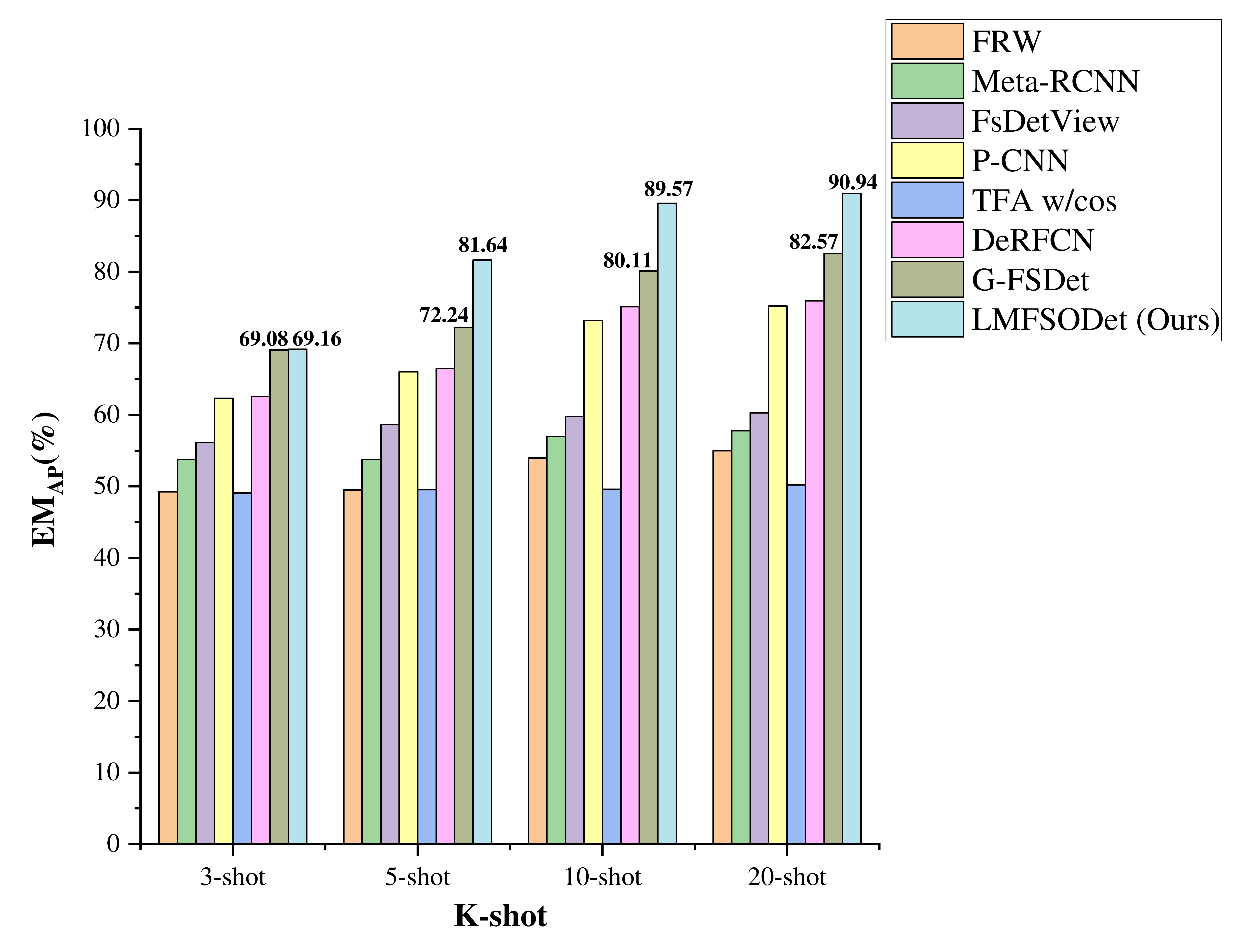}\\
		\caption{Schematic diagram of applying ECES to the performance evaluation of advanced few-shot object detectors in NWPU VHR-10.v2 Split 1.}
		\label{NWPU_ECES}
	\end{figure}
	Our experimental setup aligns closely with the majority of networks based on meta-learning frameworks. It is noteworthy that our $K$-shot configuration is determined by the number of labels, rather than the number of images. For instance, in the 3-shot training mode, each category is limited to only 3 labels for training data.

	During both the base training and few-shot fine-tuning stages, the detector is trained using the SGD optimizer with a weight decay of 0.0005 and a momentum of 0.937. All experiments are conducted with a training batch size of 6, utilizing an NVIDIA GeForce GTX 2080Ti GPU. For inference, we set the confidence threshold and IoU threshold in non-maximum suppression (NMS) to 0.05 and 0.5, respectively. Three sets of few-shot sets, comprising both base and novel classes, are randomly constructed. We present the averaged results obtained from three repetitions of the experiments. 
	The similar protocol is followed for all other methods (G-FSDet \cite{GFSOD}, Meta-RCNN \cite{meta-rcnn}, \emph{etc}) in our experiments.
	\subsection{Comparing Methods}
	To showcase the efficacy of our proposed LMFSODet, we present a comparative analysis of its performance against several state-of-the-art methods on the DIOR and NWPU VHR-10.v2 datasets. The evaluated FSOD methods encompass four meta-learning approaches including FRW \cite{FRW}, Meta-RCNN \cite{meta-rcnn}, FsDetView \cite{FsD}, and P-CNN\cite{PCNN} and three transfer-learning methods including TFA \cite{TFA},  DeRFCN \cite{DeRCFN}, and G-FSDet \cite{GFSOD}.
	
	We adopt mAP and ECES jointly to measure the performance of the detectors. In our experiments, we initially use mAP provided by Eq. \textcolor{blue}{(\ref{MAP})}, a metric widely adopted by previous scholars, as a measurement standard. Subsequently, we employ ECES provided by Eq. \textcolor{blue}{(\ref{ECES})} to assess the performance of advanced detectors.
	Following Eq. \textcolor{blue}{(\ref{ECES_weight})}, in the NWPU VHR-10.v2 dataset, there should be 
	\begin{equation}
		W_B=3,W_N=7,
	\end{equation}
	While in DIOR dataset, there should be
	\begin{equation}
		W_B=1,W_N=3.
	\end{equation}
	\subsection{Results on NWPU VHR-10.v2}
	\subsubsection{Results Based on mAP}
	Table \ref{tab3} provides a comprehensive overview of the performance of leading FSOD models, assessed in terms of mAP on the NWPU VHR-10.v2 dataset, alongside the results obtained by our proposed model. In the table, the model achieving the highest performance is highlighted in red and bold, while the second-best performance is indicated in blue and bold.
	
	Examining the results, TFA \cite{TFA} consistently maintains its superiority, demonstrating notably high accuracy across base classes in 3-shot, 5-shot, and 10-shot scenarios. The recently introduced G-FSDet \cite{GFSOD} exhibits significant advancements in novel class detection, contributing to an admirable overall accuracy. Our LMFSODet model, particularly in the 3-shot scenario, split 1, outperforms several advanced detectors in novel class detection accuracy. Notably, in the same scenario split 2, our base class detection accuracy even surpasses that of TFA \cite{TFA}.
	Specifically, in the 3-shot scenario, our LMFSODet demonstrates a marginally inferior performance compared to G-FSDet \cite{GFSOD}. We attribute this variance to G-FSDet's \cite{GFSOD} utilization of metric-based effective feature extraction, a technique that emphasizes the extraction of proficient features. Moreover, the influence of lack of samples (including $\mathcal{N}_{meta}$ and $\mathcal{P}_{meta}$ ) to be used in the 3-shot scenario also matters to the result.
	
	In the 5-shot scenario, our model excels in novel class detection accuracy in split 1, trailing marginally behind G-FSDet \cite{GFSOD} in split 2. Remarkably, in subsequent 10-shot and 20-shot scenarios, LMFSODet demonstrates remarkable superiority, achieving peak metrics and outperforming the second-best model by a considerable margin. 
	This observation implies that as the number of training labels increases, there is a corresponding augmentation in the abundance of informative negative samples, meaning more $\mathcal{N}_{meta}$ can be utilized. Consequently, the discernible effects of our meta-sampling and meta-cross loss mechanisms become more pronounced.
	
	Moreover, within the 10-shot scenario, Fig. \ref{NWPU_VISU} presents visual comparison results between LMFSODet and G-FSDet \cite{GFSOD} on the NWPU VHR-10.v2 dataset. This visualization provides supplementary insights into the relative performance of the two models, thereby reinforcing the effectiveness of LMFSODet in real-world scenarios.
	
	In summary, these findings suggest that our proposed detector excels not only in recognizing novel classes but also adeptly handles base classes without significant degradation in performance, a crucial attribute for a balanced and effective detector.
	
	\subsubsection{Results Based on ECES}
	In mainstream $mAP$ evaluation criteria, our detector's performance demonstrates superior capabilities compared to current state-of-the-art detectors. Now we reevaluate the performance of all advanced detectors employing ECES, and the result is shown in Fig. \ref{NWPU_ECES}.
	It is evident that our proposed detector exhibits higher detection accuracy than current state-of-the-art detectors mentioned in the majority of scenarios, substantiating the outstanding overall capabilities of our detector.
	\begin{figure}[tb]
		\centering
		\includegraphics[width=3.6in, keepaspectratio]{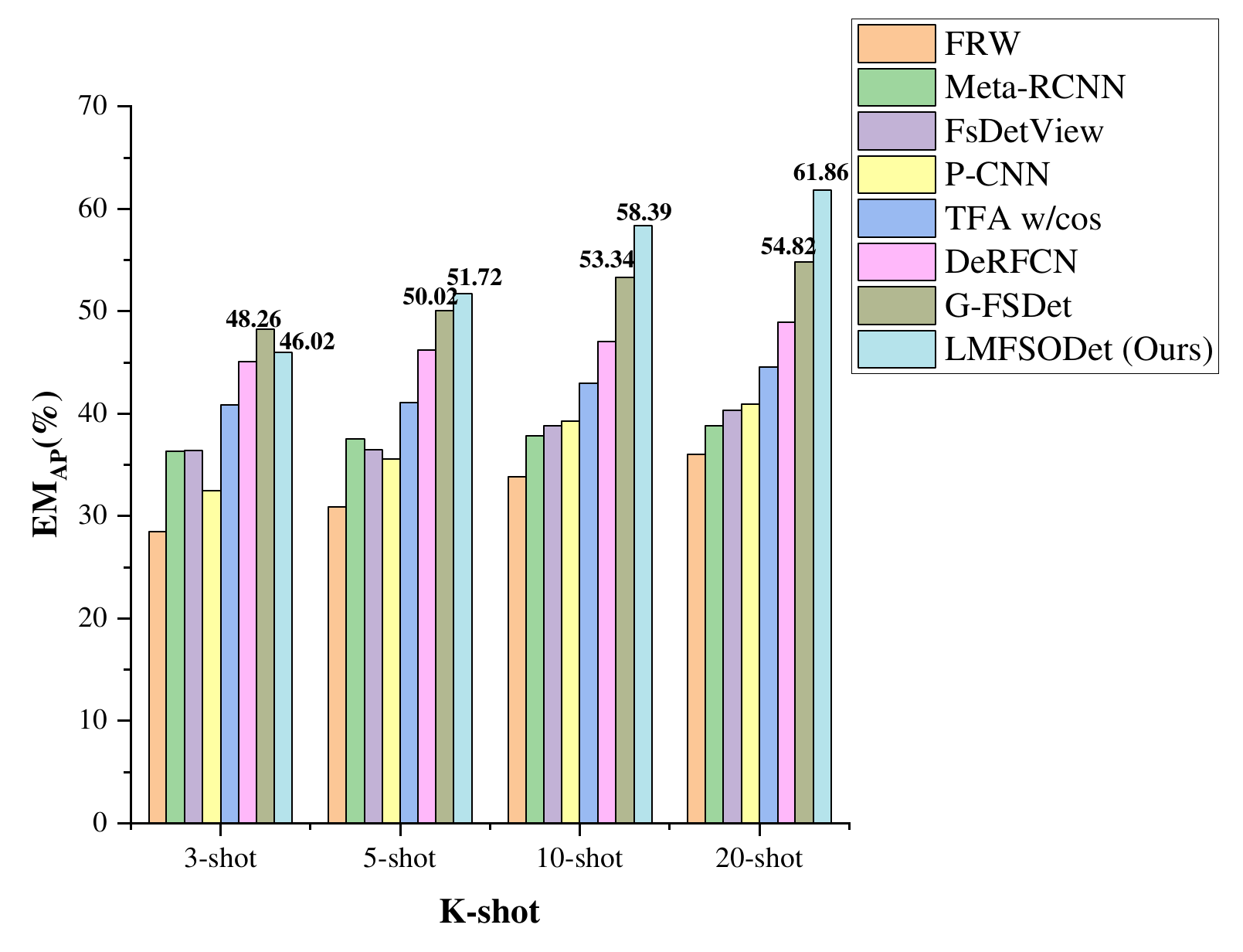}\\
		\caption{Schematic diagram of applying ECES to the performance evaluation of advanced few-shot object detectors in DIOR Split 1.}
		\label{DIOR_ECES}
	\end{figure}
	\begin{table*}[htb] 	
		\caption{FSOD models performance on the DIOR test set across various scenarios, including 3-shot, 5-shot, 10-shot, and 20-shot settings. The optimal outcomes are marked in red, while the second-best results are represented in blue. All data results are obtained by averaging the results of three independent experiments.
		}
		\label{tab4}
		\renewcommand{\arraystretch}{1.2}
		\centering
		\begin{tabular}{cc|ccc|ccc|ccc|ccc} 	
			\toprule  
			Split & Method & & 3-shot  &  & &  5-shot &  &  &10-shot  &  &  &20-shot  &  \\
			\midrule
			&  & Base & Novel & All & Base & Novel & All & Base & Novel & All & Base & Novel & All \\
			
			\multirow{8}*{1} & FRW \cite{FRW}& 49.4 & 7.5 & 38.9 & 49.7 & 12.1 & 40.3 & 49.5 & 18.1 & 41.7 & 50.0 & 22.0 & 43.0 \\
			
			& Meta-RCNN \cite{meta-rcnn} & 60.62 & 12.02 & 48.47 & 62.01 & 13.09 & 49.78 & 61.55 & 14.07 & 49.68 & 63.21 & 14.45 & 51.02 \\
			
			& FsDetView \cite{FsD} & 59.54 & 13.19 & 47.95 & 58.58 & 14.29 & 47.51 & 59.64 & 18.02 & 49.24 & 62.69 & 18.01 & 51.52 \\
			
			& P-CNN \cite{PCNN}& 47.0 & 18.0 & 39.8 & 48.4 & 22.8 & 42.0 & 50.9 & 27.6 & 45.1 & 52.2 & 29.6 & 46.8 \\
			
			& TFA w/cos \cite{TFA}& \textcolor{blue}{\textbf{70.32}} & 11.35 & 55.58 & \textcolor{blue}{\textbf{70.51}} & 11.57 & 55.78 & 70.52 & 15.37 & 56.73 & 71.07 & 17.96 & 57.79 \\
			
			& DeRFCN \cite{DeRCFN}& 61.90 & \textcolor{red}{\textbf{28.25}} & 53.48 & 62.04 & \textcolor{blue}{\textbf{30.30}}  & 54.11 & 61.51 & 32.64 & 54.29 & 62.44 & 35.37 & 55.67 \\
			
			& G-FSDet \cite{GFSOD}& 68.94 & \textcolor{blue}{\textbf{27.57}} &\textcolor{blue}{\textbf{58.61}} & 69.52 & \textcolor{red}{\textbf{30.52}}  &  \textcolor{blue}{\textbf{59.72}} & \textcolor{blue}{\textbf{69.03}} & \textcolor{blue}{\textbf{37.64}} & \textcolor{blue}{\textbf{61.16}} & \textcolor{blue}{\textbf{69.80}} & \textcolor{blue}{\textbf{39.83}} & \textcolor{blue}{\textbf{62.31}} \\
			
			& Ours & \textcolor{red}{\textbf{72.30}}  & 19.73 & \textcolor{red}{\textbf{59.16}} & \textcolor{red}{\textbf{75.44}} & 28.00 &  \textcolor{red}{\textbf{63.58}} & \textcolor{red}{\textbf{77.22}} & \textcolor{red}{\textbf{39.56}} &  \textcolor{red}{\textbf{67.80}} & \textcolor{red}{\textbf{78.53}}& \textcolor{red}{\textbf{45.18}}&\textcolor{red}{\textbf{70.19}} \\
			\midrule
			\multirow{8}*{2} & FRW \cite{FRW}& 48.5 & 4.8 & 37.6 & 46.8 & 7.0 & 36.9 & 46.4 & 9.0 & 37.1 & 43.5 & 14.1 & 36.2 \\
			
			& Meta-RCNN \cite{meta-rcnn}& 62.55 & 8.84 & 49.12 & 63.14 & 10.88 & 50.07 & 63.28 & 14.90 & 51.18 & 63.86 & 16.71 & 52.07 \\
			
			& FsDetView \cite{FsD}& 58.88 & 10.83 & 46.87 & 60.31 & 9.63 & 47.64 & 61.16 & 13.57 & 49.26 & 61.16 & 14.76 & 49.56 \\
			
			& P-CNN \cite{PCNN}& 48.9 & \textcolor{blue}{\textbf{14.5}} & 40.3 & 49.1 & 14.9 & 40.6 & 52.5 & 18.9 & 44.1 & 51.6 & \textcolor{blue}{\textbf{22.8}} & 44.4 \\
			
			& TFA w/cos \cite{TFA}& \textcolor{red}{\textbf{70.75}} & 5.77 & 54.51 &  \textcolor{blue}{\textbf{70.79}} & 8.19 & 55.14 &\textcolor{blue}{\textbf{69.63}} & 8.71 & 54.63 & \textcolor{blue}{\textbf{70.02}} & 12.18 & 55.56 \\
			
			& DeRFCN \cite{DeRCFN}& 61.41 & \textcolor{red}{\textbf{14.55}} & 49.70 & 61.79 &  \textcolor{red}{\textbf{16.45}} & 50.45 & 61.42 & 18.4 & 50.69 & 62.79 & 21.13 & 52.38 \\
			
			& G-FSDet \cite{GFSOD}& 69.20 & 14.13 & \textcolor{red}{\textbf{55.43}} & 69.25 & 15.84 & \textcolor{blue}{\textbf{55.87}} & 68.71 & \textcolor{blue}{\textbf{20.70}} & \textcolor{blue}{\textbf{56.70}} & 68.18 & 22.69 & \textcolor{blue}{\textbf{56.86}} \\
			
			& Ours &\textcolor{blue}{\textbf{69.94}} & 9.49 & \textcolor{blue}{\textbf{54.83}} &  \textcolor{red}{\textbf{74.28}} &\textcolor{blue}{\textbf{16.41}}&\textcolor{red}{\textbf{59.81}} & \textcolor{red}{\textbf{78.37}} &\textcolor{red}{\textbf{31.92}}& \textcolor{red}{\textbf{66.75}} & \textcolor{red}{\textbf{79.95}} &\textcolor{red}{\textbf{42.67}} & \textcolor{red}{\textbf{70.63}} \\
			\midrule
			\multirow{8}*{3} & FRW \cite{FRW}& 45.5 & 7.8 & 36.1 & 47.9 & 13.7 & 39.3 & 44.5 & 13.8 & 36.8 & 43.5 & 18.5 & 37.3 \\
			
			& Meta-RCNN \cite{meta-rcnn}& 61.93 & 9.10 & 48.72 & 63.44 & 12.29 & 50.66 & 62.57 & 11.96 & 49.92 & 65.53 & 16.14 & 53.18 \\
			
			& FsDetView \cite{FsD}& 61.00 & 7.49 & 47.63 & 61.33 & 12.61 & 49.15 & 61.94 & 11.49 & 49.32 & 65.17 & 17.02 & 53.14 \\
			
			& P-CNN \cite{PCNN}& 49.5 & \textcolor{red}{\textbf{16.5}} & 41.3 & 49.9 & \textcolor{blue}{\textbf{18.8}}  & 42.1 & 52.1 & 23.3 & 44.9 & 53.1 & 28.8 & 47.0 \\
			
			& TFA w/cos \cite{TFA}& \textcolor{red}{\textbf{71.95}} & 8.36 & \textcolor{blue}{\textbf{56.05}} & \textcolor{blue}{\textbf{71.64}}  & 10.13 & 56.26 &\textcolor{blue}{\textbf{72.56}} & 10.75 & 57.11 & \textcolor{blue}{\textbf{73.13}} & 17.99 & 59.35 \\
			
			& DeRFCN \cite{DeRCFN}& 63.56 & 15.78 & 51.47 & 63.25 & 18.73 & 52.11 & 64.55 & 20.43 & 53.52 & 64.56 & 25.13 & 54.71 \\
			
			& G-FSDet \cite{GFSOD}&71.10 & \textcolor{blue}{\textbf{16.03}} & \textcolor{red}{\textbf{57.34}} & 70.18 & \textcolor{red}{\textbf{23.25}}  & \textcolor{blue}{\textbf{58.43}}  & 71.08 & \textcolor{blue}{\textbf{26.24}}  & \textcolor{blue}{\textbf{59.87}}  & 71.26 & \textcolor{blue}{\textbf{32.05}} & \textcolor{blue}{\textbf{61.46}} \\
			
			& Ours & \textcolor{blue}{\textbf{71.12}} & 10.17 & 55.88 &\textcolor{red}{\textbf{76.65}}  & 13.59 & \textcolor{red}{\textbf{60.88}}  &\textcolor{red}{\textbf{77.94}}  & \textcolor{red}{\textbf{29.24}}  & \textcolor{red}{\textbf{65.76}}  & \textcolor{red}{\textbf{80.13}} &\textcolor{red}{\textbf{34.77}} & \textcolor{red}{\textbf{68.79}} \\
			\midrule
			\multirow{8}*{4} & FRW \cite{FRW}& 48.2 & 3.7 & 37.1 & 48.5 & 6.8 & 38.1 & 45.7 & 7.2 & 36.1 & 44.4 & 12.2 & 36.4 \\
			
			& Meta-RCNN \cite{meta-rcnn}& 61.73 & 13.94 & 49.78 & 62.60 & 15.84 & 50.91 & 62.23 & 15.07 & 50.44 & 63.24 & 18.17 & 51.98 \\
			
			& FsDetView \cite{FsD}& 58.90 & 14.28 & 47.75 & 58.97 & 15.95 & 48.22 & 60.37 & 15.37 & 49.12 & 60.89 & 16.96 & 49.91 \\
			
			& P-CNN \cite{PCNN}& 49.8 & \textcolor{blue}{\textbf{15.2}} & 41.2 & 49.9 & 17.5 & 41.8 & 51.7 & 18.9 & 43.5 & 52.3 & 25.7 & 45.7 \\
			
			& TFA w/cos \cite{TFA}& \textcolor{blue}{\textbf{68.57}} & 10.42 & 54.03 & \textcolor{blue}{\textbf{68.85}} & 14.29 & 55.21 & \textcolor{blue}{\textbf{68.58}} & 14.35 & 55.03 & \textcolor{blue}{\textbf{68.86}} & 12.01 & 54.65 \\
			
			& DeRFCN \cite{DeRCFN}& 59.81 & 10.83 & 47.50 & 59.71 & 18.62 & 49.44 & 59.54 & 21.61 & 49.99 & 60.09 & 27.61 & 51.97 \\
			
			& G-FSDet \cite{GFSOD}&\textcolor{red}{\textbf{69.01}} & \textcolor{red}{\textbf{16.74}} & \textcolor{red}{\textbf{55.95}} & 67.96 & \textcolor{blue}{\textbf{21.03}} & \textcolor{blue}{\textbf{56.30}} & 68.55 & \textcolor{blue}{\textbf{25.84}}  &\textcolor{blue}{\textbf{57.87}}  & 67.73 & \textcolor{blue}{\textbf{31.78}} & \textcolor{blue}{\textbf{58.75}} \\
			
			& Ours & 68.35 & 12.09 & \textcolor{blue}{\textbf{54.28}} & \textcolor{red}{\textbf{74.39}} & \textcolor{red}{\textbf{25.89}} & \textcolor{red}{\textbf{62.26}} & \textcolor{red}{\textbf{78.83}} & \textcolor{red}{\textbf{30.94}} & \textcolor{red}{\textbf{66.86}}& \textcolor{red}{\textbf{80.62}} & \textcolor{red}{\textbf{39.68}}& \textcolor{red}{\textbf{70.39}} \\
			\bottomrule
		\end{tabular}
	\end{table*}
	\begin{figure*}[htb]
		\centering
		
		\subfloat[Results of G-FSDet \cite{GFSOD} on the DIOR dataset in the 10-shot scenario.]{\includegraphics[width=7.0in]{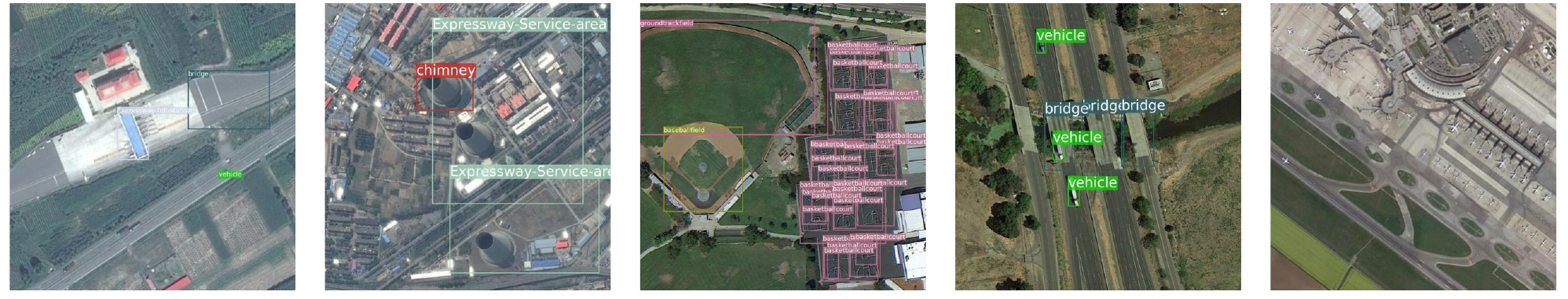} }
		
		\subfloat[Results of LMFSODet (ours) on the DIOR dataset in the 10-shot scenario.]{\includegraphics[width=7.0in]{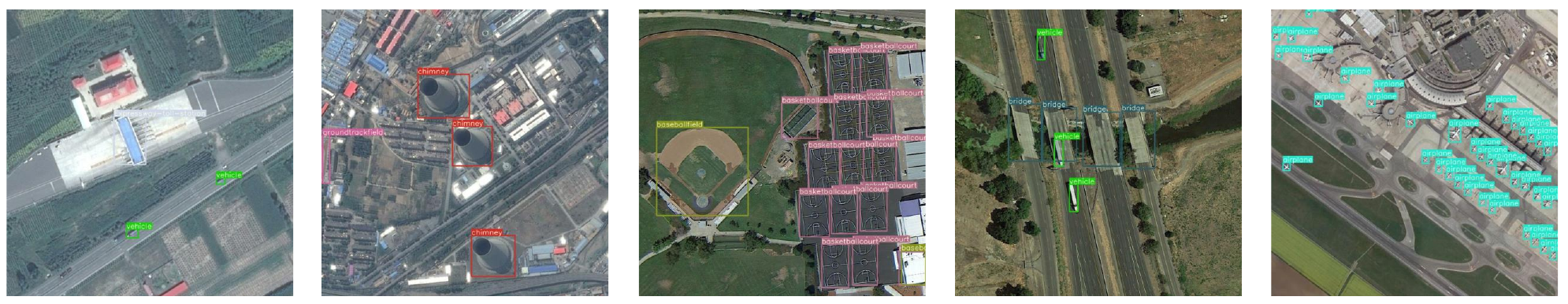} }
		\caption{Visualization results of LMFSODet (ours) compared to G-FSDet \cite{GFSOD} on the DIOR dataset in the 10-shot scenario.}
		\label{DIOR_V}
	\end{figure*}
	\subsection{Results on DIOR}
	\subsubsection{Results Based on mAP}
	Table \ref{tab4} clearly shows the performance of our detector and various state-of-the-art detectors on the DIOR dataset. The training strategy of TFA \cite{TFA}, which emphasizes base class training, makes it highly ideal for preserving base class knowledge, but it also limits the improvement of its accuracy in new class detection. G-FSDet \cite{GFSOD}, with its more comprehensive training strategy and compensation module, achieves further improvement, performing well in the 3-shot scenario. In contrast, our proposed LMFSODet, with a novel detection approach and loss strategy, enhances the detection ability of new classes while preserving base class knowledge in most scenarios. 
	It can be observed that under the 5-shot setting, our detection accuracy surpasses all advanced detectors mentioned in most metrics, and it demonstrates overwhelming advantages in subsequent 10-shot and 20-shot scenarios. However, in the 3-shot scenario, our detection accuracy is slightly lower than G-FSDet \cite{GFSOD}, partly due to its superior training strategy and module design in the meta-learning feature obtaining strategy and insufficient samples to be delved by meta-sampling, which are discussed before. 
	
	In the 10-shot scenario, the visualization results on DIOR dataset comparing LMFSODet with G-FSDet \cite{GFSOD} are depicted in Fig. \ref{DIOR_V}. 
	It can be observed that G-FSDet \cite{GFSOD} exhibits detection omissions for objects with larger scale range, which is one of the issues stemming from the two-stage detectors. Although many few-shot detectors based on two-stage detectors are employing various methods to address the multiscale problem, such as using FPN to alleviate scale variations as mentioned in \cite{ML1}, we believe that the inherent advantage of one-stage detectors with natural multiscale training and ultra-fast detection speed aligns better with the practical requirements of real-world systems. And one thing to notice is that
	our primary focus lies in the optimization of leveraging the extensive samples generated through meta-learning.
	
	\subsubsection{Results Based on ECES}
	Similar to NWPU VHR-10.v2 dataset, we also evaluate the performance of detectors on the DIOR dataset based on ECES, as shown in Fig. \ref{DIOR_ECES}.
	From the perspective of ECES, our detector still exhibits high detection accuracy. Except for a slightly lower performance than G-FSDet \cite{GFSOD} in the 3-shot scenario, LMFSODet outperforms the mentioned advanced detectors. 
	\subsection{Ablation Analysis}
	\subsubsection{Analysis Based on Baseline}
	To assess the advancement of all designed modules and strategies, we conducted ablation experiments. The subjects of these experiments included meta-training, encompassing meta-sampling and the meta-detection head (with DMP), as well as the meta-cross loss. Given the complementary nature of meta-sampling and DMP, they were concomitantly subjected to ablation within the context of meta-training. Conversely, while the meta-cross loss is founded on meta-sampling, it can be independently ablated as a distinct module to substantiate its efficacy. The results based on mAP are shown in Table \ref{tab7}.
	\begin{figure*}[htb]
		\centering
		
		\subfloat[Results of YOLOV7 \cite{yolov7} on the DIOR and NWPU VHR-10.v2 dataset in the 10-shot scenario.]{\includegraphics[width=7.0in]{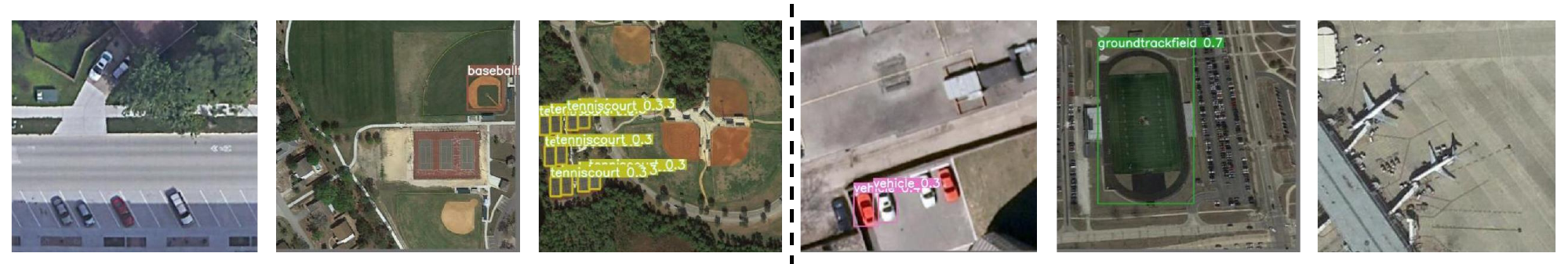} }
		
		\subfloat[Results of LMFSODet (ours) on the DIOR and NWPU VHR-10.v2 dataset in the 10-shot scenario.]{\includegraphics[width=7.0in]{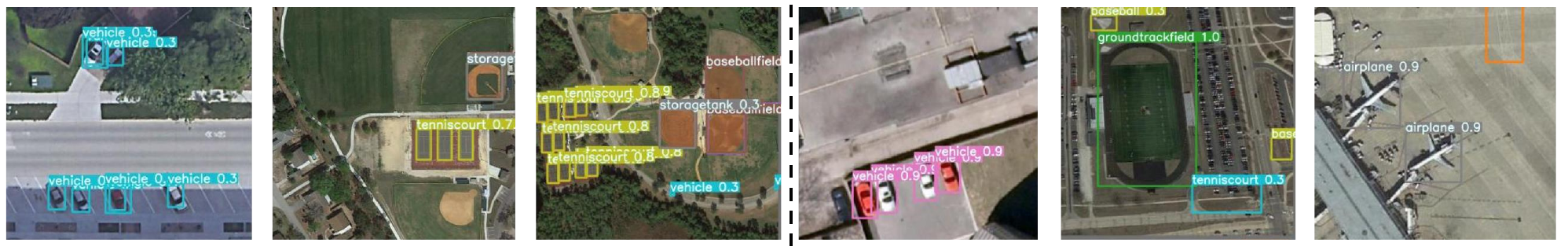} }
		\caption{Visualization results of LMFSODet (ours) compared to YOLOv7 \cite{yolov7} (baseline) on the DIOR and NWPU VHR-10.v2 dataset in the 10-shot scenario. 
			The left three figures depict the results of the DIOR dataset, while the right three figures represent the results of the NWPU VHR-10.v2 dataset.}
		\label{AB_F}
	\end{figure*}
	\begin{table*}[htb] 	
		\caption{The results of the ablation experiments on meta-training and meta-cross loss. ``All" entails utilizing meta-sampling and meta-cross loss, where the weight $\lambda$ is endowed with an appropriate value.
		}
		\label{tab7}
		\renewcommand{\arraystretch}{1.2}
		\centering
		\begin{tabular}{cc|ccc|ccc|ccc|ccc} 	
			\toprule  
			Dataset & Method &  &3-shot  &  &&  5-shot  &  & &10-shot   &  & &  20-shot &  \\
			\midrule
			&  & Base & Novel & All & Base & Novel & All & Base & Novel & All & Base & Novel & All \\
			
			\multirow{3}*{DIOR} & YOLOv7 & 56.34 & 11.27 & 45.07 & 61.59 & 24.44 & 52.30 &60.17 & \textcolor{blue}{\textbf{38.67}} & 54.79 &58.38 &\textcolor{red}{\textbf{50.39}} & 56.38 \\
			
			& Meta-training & \textcolor{blue}{\textbf{67.29}} & \textcolor{red}{\textbf{20.84}} &\textcolor{blue}{\textbf{55.68}} & \textcolor{red}{\textbf{76.68}} &\textcolor{blue}{\textbf{26.31}}  &\textcolor{red}{\textbf{64.09}} &\textcolor{red}{\textbf{78.51}}  & 36.25 &\textcolor{red}{\textbf{67.94}} &\textcolor{red}{\textbf{79.75}} &{42.12}& \textcolor{red}{\textbf{70.34}}\\
			
			& All &\textcolor{red}{\textbf{72.30}} & \textcolor{blue}{\textbf{19.73}} & \textcolor{red}{\textbf{59.16}} &  \textcolor{blue}{\textbf{75.44}} &\textcolor{red}{\textbf{28.0}}&\textcolor{blue}{\textbf{63.58}} & \textcolor{blue}{\textbf{77.22}} &\textcolor{red}{\textbf{39.56}}& \textcolor{blue}{\textbf{67.80}} & \textcolor{blue}{\textbf{78.53}} &\textcolor{blue}{\textbf{45.18}} & \textcolor{blue}{\textbf{70.19}} \\
			\midrule
			\multirow{3}*{NWPU.v2} & YOLOv7 & 84.00 & 31.93& 68.38 & 86.58 & 45.77 & 74.34& \textcolor{blue}{\textbf{87.23}}& 68.38& 81.57 & 82.45 & 77.13 & 80.85\\
			
			& Meta-training &\textcolor{red}{\textbf{89.54}} & \textcolor{blue}{\textbf{41.10}}&\textcolor{blue}{\textbf{75.01}} &\textcolor{blue}{\textbf{90.37}} &\textcolor{blue}{\textbf{65.49}} & \textcolor{blue}{\textbf{82.91}} & 87.01 & \textcolor{blue}{\textbf{82.84}} & \textcolor{blue}{\textbf{85.76}} &\textcolor{blue}{\textbf{89.78}} &\textcolor{blue}{\textbf{86.36}} & \textcolor{blue}{\textbf{88.75}}\\
			
			& All  & \textcolor{blue}{\textbf{85.15}} & \textcolor{red}{\textbf{53.16}} & \textcolor{red}{\textbf{75.55}} & \textcolor{red}{\textbf{90.62}} & \textcolor{red}{\textbf{72.65}} & \textcolor{red}{\textbf{85.23}} & \textcolor{red}{\textbf{91.16}}& \textcolor{red}{\textbf{87.97}} & \textcolor{red}{\textbf{90.20}}& \textcolor{red}{\textbf{91.67}} 
			& \textcolor{red}{\textbf{90.20}}  & \textcolor{red}{\textbf{91.23}}  \\
			\bottomrule
		\end{tabular}
	\end{table*}
	As depicted in Table \ref{tab7},  YOLOv7, employed as a standard object detector, inherently exhibits commendable detection performance. Nonetheless, it faces challenges in effectively handling few-shot object detection tasks. Upon the incorporation of our designed meta-training, the detector demonstrates enhanced adaptability to few-shot detection tasks, leading to an average accuracy improvement of around 10\%, revealing the effectiveness of our proposed meta-training mechanism.
	The visualization results of the baseline and proposed detector on the two major remote sensing datasets are depicted in Fig. \ref{AB_F}.
	
	Additionally, according to Table \ref{tab7},
	the meta-cross loss aids the detector in extracting knowledge from informative negative samples. This valuable knowledge can stem from negative samples belonging to either base or novel classes, thereby making the improvement in accuracy contingent upon the categories that provide useful knowledge. A higher prevalence of negative samples from base classes leads to a more significant enhancement in accuracy for base classes, while a greater abundance of negative samples from novel classes results in a more pronounced improvement in accuracy for novel classes.
	
\begin{figure}[htb]
		\centering
		\includegraphics[width=3.5in, keepaspectratio]{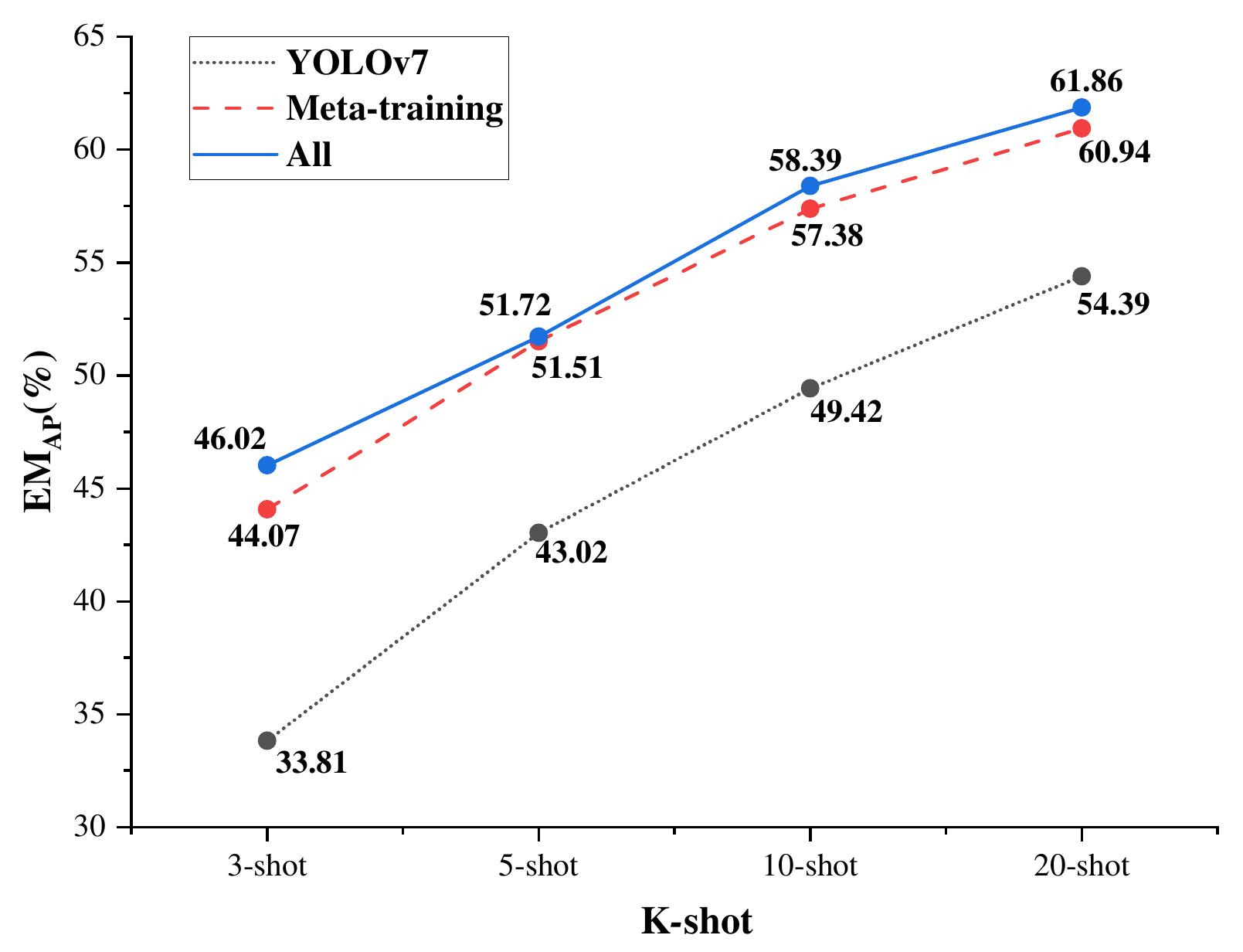}\\
		\caption{ Modules ablation experiments results based on ECES in DIOR dataset split 1. ``All" entails utilizing meta-sampling and meta-cross loss, where the weight $\lambda$ is endowed with an appropriate value.}
		\label{DIOR_AB}
	\end{figure}

\begin{figure}[htb]
		\centering
		\includegraphics[width=3.5in, keepaspectratio]{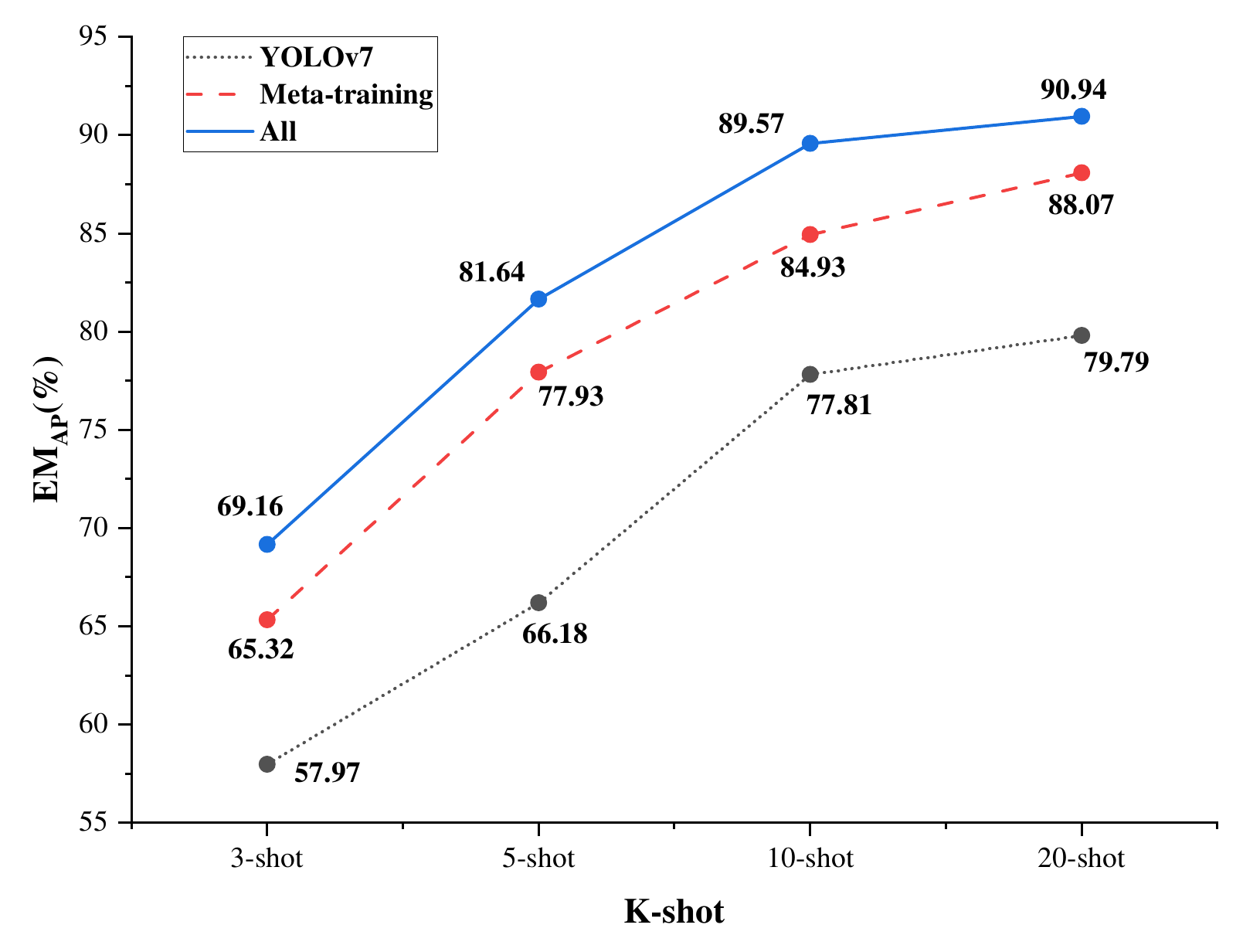}\\
		\caption{Modules ablation experiments results based on ECES in NWPU VHR-10.v2 dataset split 1. ``All" entails utilizing meta-sampling and meta-cross loss, where the weight $\lambda$ is endowed with an appropriate value.}
		\label{NWPU_AB}
	\end{figure}

	We also apply ECES to reevaluate the corresponding ablation experiment results, as shown in Fig. \ref{DIOR_AB} and Fig. \ref{NWPU_AB}.
	Obviously, our detector shows a significant improvement over the baseline.
	Moreover, with the value of $K$ increasing, its enhancement effect becomes more pronounced. The reason lies in the fact that with smaller $K$ values, the number of negative samples decreases, resulting in fewer valuable negative samples and thus less useful information extraction. Therefore, the improvement in the 3-shot scenario is not substantial, but as the sample quantity increases, the effectiveness of meta-cross loss tends to become more prominent.
	\subsubsection{Analysis Based on Meta-RCNN}
\begin{figure}[htb]
		\centering
		\includegraphics[width=3.5in, keepaspectratio]{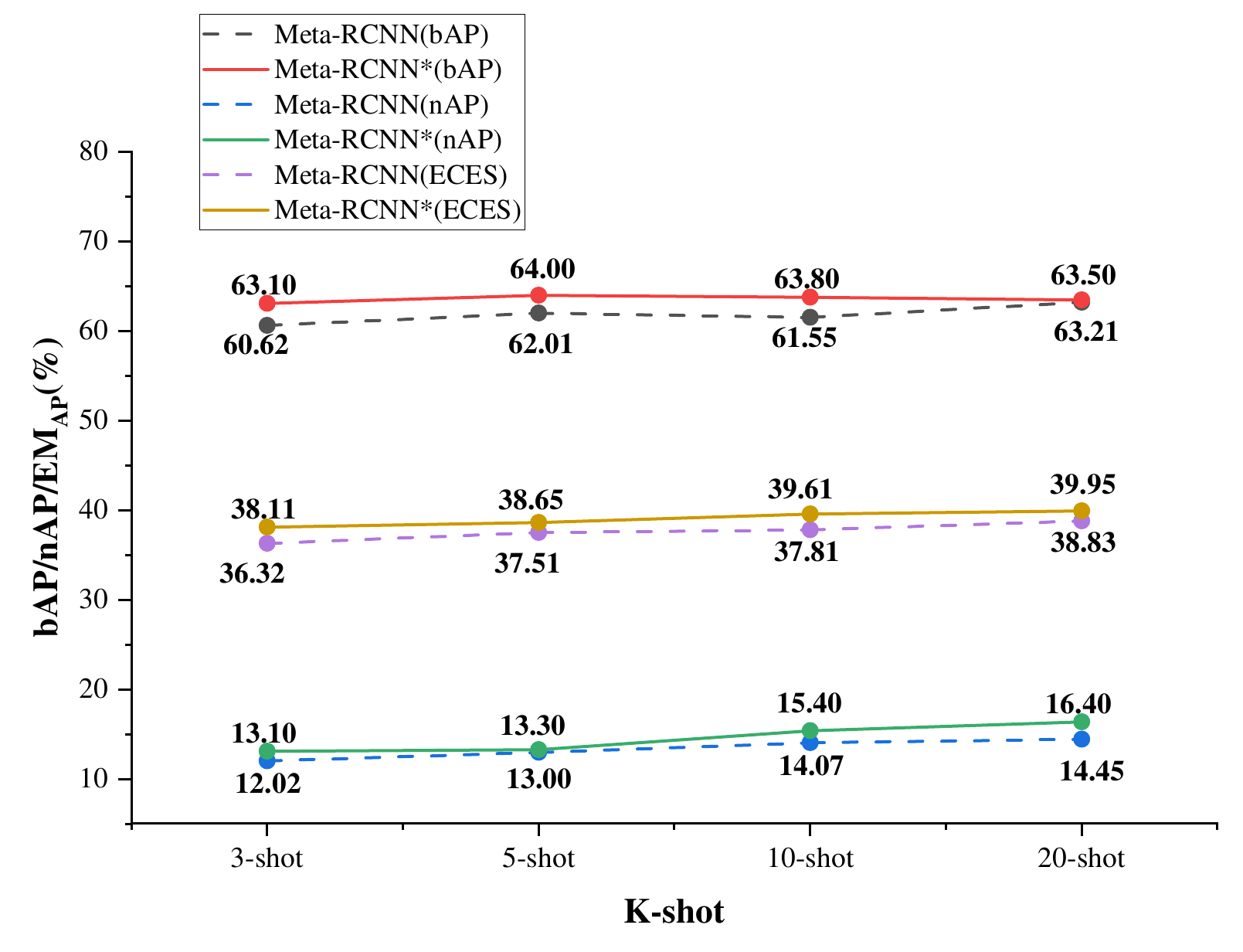}\\
		\caption{Schematic diagram of the ablation experimental results based on Meta-RCNN on DIOR split 1 dataset. We evaluated the results using two evaluation metrics, mAP and ECES. Meta-RCNN represents the direct experimental results from \cite{meta-rcnn}, while Meta-RCNN* indicates the incorporation of our designed meta-sampling and meta-cross loss on top of Meta-RCNN \cite{meta-rcnn}.}
		\label{Meta-RCNN_AB}
	\end{figure}

	In order to substantiate the efficacy of our module in refining meta-learning strategies, we conducted additional experiments on Meta-RCNN \cite{meta-rcnn} by incorporating meta-sampling technique and meta-cross loss. The experimental findings are presented in Fig. \ref{Meta-RCNN_AB}.
	As depicted in Fig. \ref{Meta-RCNN_AB}, the integration of our proposed strategy and module results in an approximate 2\% enhancement in the comprehensive detection accuracy of Meta-RCNN \cite{meta-rcnn} across diverse scenarios. This substantiates the refinement of our approach in meta-learning strategies, indicating a potential for effective transferability to few-shot object detectors based on meta-learning.
	\subsection{Complexity Comparison}
	\begin{figure}[htb]
		\centering
		\includegraphics[width=3.5in, keepaspectratio]{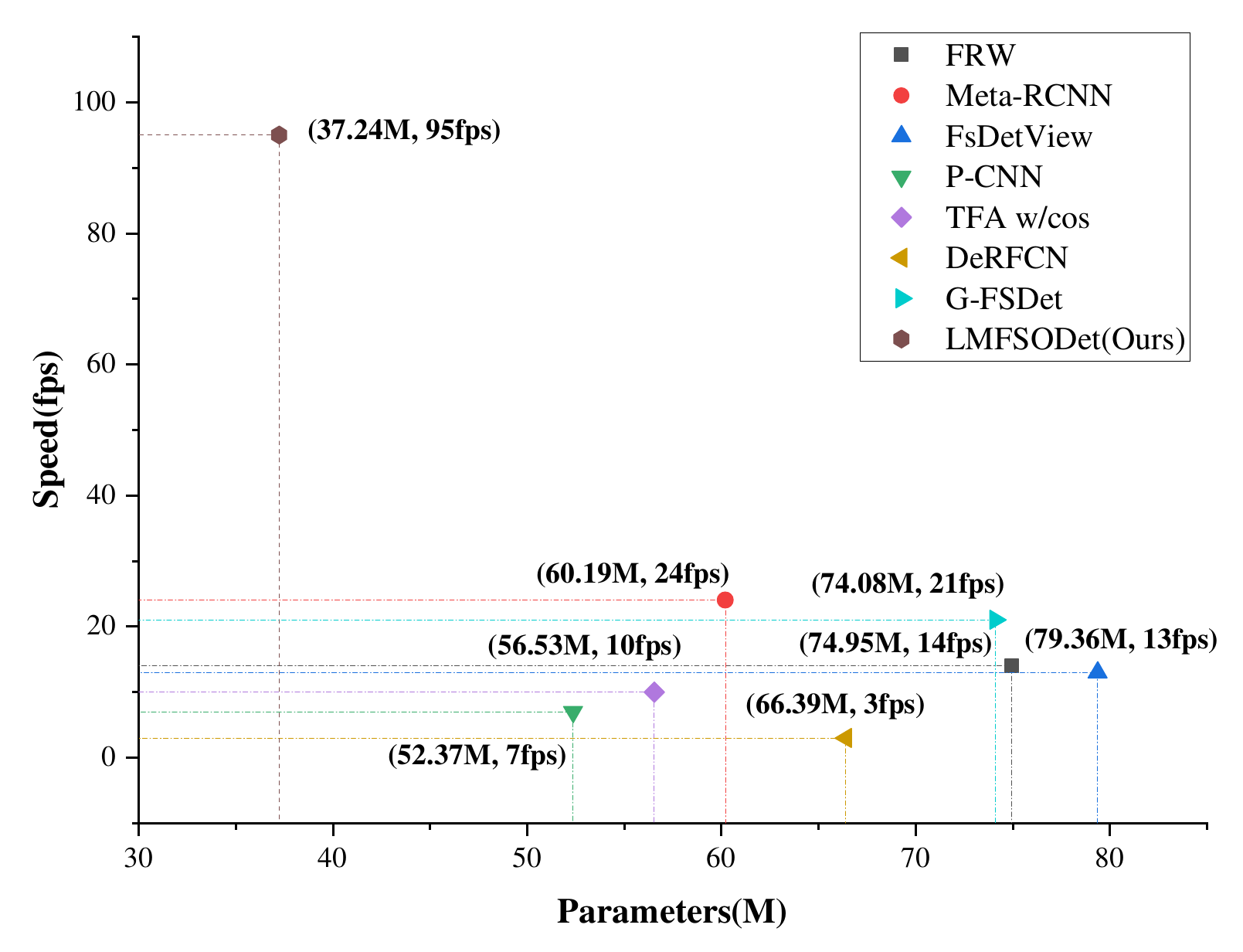}\\
		\caption{Model parameters and detection speed comparison scatter plot.}
		\label{Complexity}
	\end{figure}
	
	We performed an evaluation of model parameter sizes for the mentioned detectors, as depicted in Fig. \ref{Complexity}. 
	It is evident from the evaluation that the parameter count of our proposed detector is the smallest among various advanced detectors. Notably, this parameter metric pertains to the parameters utilized during the training phase. 
	During the prediction phase, the parameters of the support set feature extraction module can be disregarded, allowing for a further reduction in the parameter count.
	Regarding detection speed, our detector achieves an impressive rate of 95 frames per second, surpassing all other detectors, in perfect alignment with our original design intent emphasizing lightweight characteristics.
	
	\begin{table*}[tb] 	
		\caption{Analysis of Deconstruction Experiment with $\lambda$ Weight Assignment on DIOR dataset.
		}
		\label{tab6}
		\renewcommand{\arraystretch}{1.2}
		\centering
		\begin{tabular}{c|ccc|ccc|ccc|ccc}
			\midrule
			$\lambda$ & &  3-shot &  &  & 5-shot &  &  & 10-shot &  & & 20-shot  &  \\
			\midrule
			& Base & Novel & All & Base & Novel & All & Base & Novel & All & Base & Novel & All \\
			
			0 & 67.29 & 20.84 & 55.68 & 76.68 & 26.31 & 64.09 & 78.51 & 36.25 & 67.94 & 79.75 & 42.12 & 70.34 \\
			
			0.5 & 9.44 & 0.17 & 7.12 & 22.45 & 0.41 & 16.94 & 42.31 & 17.36 & 36.07 & 63.92 & 30.27 & 55.51 \\
			
			0.1 & 45.26 & 16.30 & 38.02 & 62.79 & 25.39 & 53.44 & 77.22 & 39.56 & 67.80 & 78.53 & 45.18 & 70.19 \\
			
			0.01 & 65.68 & 19.02 & 54.01 & 72.28 & 25.47 & 60.58 & 75.07 & 35.78 & 65.25 & 78.49 & 41.91 & 69.35 \\
			
			0.005 & 69.03 & 17.99 & 56.27 & 75.44 & 28.00 & 63.58 & 79.98 & 38.23 & 69.54 & 78.46 & 42.49 & 69.47 \\
			
			0.001 & 72.30 & 19.73 & 59.16 & 77.70 & 25.95 & 64.76 & 80.41 & 36.15 & 69.34 & 79.16 & 41.29 & 69.70 \\
			\bottomrule
		\end{tabular}
	\end{table*}
	\begin{table*}[tb] 	
		\caption{Analysis of Deconstruction Experiment with $\lambda$ Weight Assignment on NWPU VHR-10.v2 dataset.
		}
		\label{tab8}
		\renewcommand{\arraystretch}{1.2}
		\centering
		\begin{tabular}{c|ccc|ccc|ccc|ccc}
			\midrule
			$\lambda$ &  & 3-shot &  & & 5-shot  &  &  & 10-shot &  &  & 20-shot &  \\
			\midrule
			& Base & Novel & All & Base & Novel & All & Base & Novel & All & Base & Novel & All \\
			
			0 & 89.54 & 41.10 & 75.01 & 90.37 & 65.49 & 82.91 & 87.01 & 82.84 & 85.76 & 89.78 & 86.36 & 88.75\\
			
			0.5 &57.05 & 24.04 & 47.14 & 80.63 & 47.89 & 70.81 & 90.34 & 80.87 & 87.50 & 90.94 & 84.75 & 89.08 \\
			
			0.1 & 85.15 & 53.16 & 75.55 & 90.62 & 72.65 & 85.23 & 91.16 & 87.97 & 90.20 & 91.67 & 90.20 & 91.23 \\
			
			0.01 & 86.68 & 49.70 & 75.58 & 88.99 & 70.18 & 83.35 & 89.70 & 82.64 & 87.58 & 89.92 & 84.62 & 88.33 \\
			
			0.001 & 90.34 & 39.46 & 75.08 & 90.19 & 64.75 & 82.56 & 87.97 & 83.60 & 86.66 & 90.80 & 85.49 & 89.20 \\
			\bottomrule
		\end{tabular}
	\end{table*}
	The efficiency achieved can be attributed to our consistent adherence to the principles of lightweight and efficient design across all modules. In the feature extraction module, we utilize cutting-edge convolutional networks that not only preserve gradient pathways but also enhance the effectiveness of feature extraction. Simultaneously, in meta-detection head, we integrate a lightweight detection mechanism (DMP), which not only improves inter-class discriminative capabilities but also minimizes the overall parameter count.
	Considering the challenges posed by RSIs, such as multiscale variations, high resolution, and limited samples, our detector excels in achieving an exceptionally rapid detection speed while maintaining a high level of detection accuracy.
	\subsection{Study of Meta-Cross Loss Weight $\lambda$}
	To illustrate the impact factor of meta-cross loss, we conducted the following experiments.
	We explored different values of $\lambda$ and obtained the results presented in Table \ref{tab6} and Table \ref{tab8}.
	In our investigation, we thoroughly examined multiple datasets, implemented diverse splitting methodologies, and explored a range of $K$-shot scenarios. Our analysis yielded several key observations.
	Firstly, we determined that the suitable weight values typically reside within the interval of 0.001 to 0.1.
	Additionally, we noted a consistent positive correlation between the number of shots ($K$) and the weighting factor, indicating a continuous growth pattern.
	Our examination across different datasets revealed variations in the minimum weight values. Notably, as the number of categories expanded, there was a discernible trend towards smaller minimum weights.
	
	These findings provide crucial insights into exploring the impact of datasets and partitioning methods on model performance. In various scenarios, employing appropriate $\lambda$ enables the detector to better learn meta-features.
	The reason for setting a relatively small value for $\lambda$ weight lies in that it serves as a factor influencing useful negative samples, which is comparatively smaller in impact compared to positive samples. However, disregarding it entirely would prevent us from fully leveraging the dataset and meta-learning strategies. If adopted in practice, $\lambda$ weight can be set by empirical values we provide or redesign a lightweight network to have flexible control over it.
	
	\section{conclusion}
	This article presents a novel few-shot object detector, customized for RSIs, based on YOLOv7 architecture. Acknowledging the inherent challenges in RSIs, we propose a novel efficient meta-learning paradigm.
	Firstly, we introduce the meta-sampling technique to extract useful samples in meta-learning, enhancing sample data utilization compared to previous research. Combined with our designed meta-cross loss, we further address the issue of limited sample availability. Subsequently, we propose the DMP criterion to enhance the discriminative capability of the detection head across different classes, thus resolving the challenge of small inter-class variations in RSIs.
	In a nutshell, our detector prioritizes lightweight design principles, minimizing parameter count in module design as much as possible.
	We evaluate the detector's performance based on mAP and our proposed ECES standards using two notable RSI datasets. Results indicate that our detector excels in both detection speed and accuracy.
	
	Future endeavors will focus on delving deeper into the valuable information provided by meta-samples and exploring more advanced, lightweight, and efficient meta-feature extractors.
	\bibliographystyle{IEEEtran}
	\bibliography{reference}
\end{document}